\documentclass{article}
\usepackage[preprint]{neurips_2025}

\usepackage{supertabular}
\usepackage{enumitem}

\usepackage{amsmath} 
\usepackage{amsthm}

\usepackage[hidelinks]{hyperref}
\usepackage[capitalise,noabbrev]{cleveref}

\crefname{algocf}{Algorithm}{Algorithms}
\Crefname{algocf}{Algorithm}{Algorithms}
\usepackage{graphicx}
\usepackage{url} 
\usepackage{float}
\usepackage{subcaption}
\usepackage{placeins}

\usepackage[ruled,vlined]{algorithm2e}

\title{A Comparative Simulation Study of the Fairness and Accuracy of  Predictive Policing Systems in Baltimore City}

\author{
    Samin Semsar\thanks{Equal contribution.}\\
    University of Maryland, Baltimore County \\
    \texttt{samin.semsar@umbc.edu}
    \And
    Kiran Laxmikant Prabhu\footnotemark[1]\\
    University of Maryland, Baltimore County \\
    \texttt{kiran4@umbc.edu}
    \And
    Gabriella Waters\\
    Morgan State University \\
    \texttt{gabriella.waters@morgan.edu}
    \And
    James Foulds\footnotemark[1]\\
    University of Maryland, Baltimore County \\
    \texttt{jfoulds@umbc.edu}
}

\newcommand{\jf}[1]{#1} 

\begin{document}
\maketitle
\begin{abstract}
There are ongoing discussions about predictive policing systems being unfair, for example, by exhibiting racial bias. Law enforcement in some cities, such as Los Angeles, California, and Baltimore, Maryland, initiated the integration of these systems in their decision-making process\jf{es}~\cite{BaltimoreSun,FoxBaltimore}, 
\jf{and some} of these systems were advertised as being unbiased~\cite{perry2013predictive}. However, later studies discovered the methods could also be unfair due to 
feedback loops and being trained on historically biased recorded data~\cite{lum2016predict}. 
\jf{Comparative} studies on predictive policing systems are \jf{few and are not sufficiently comprehensive.} \jf{Crucially, the relative fairness of predictive policing methods and traditional hot spots policing has not been established.} 
\jf{Moreover, the relationship between fairness and accuracy~\cite{mohler2018penalized} is complex and requires further study.}  
\jf{Furthermore, the case of Baltimore City, Maryland, USA, has not yet been systematically analyzed, despite its \jf{relevance} as an early adopter of predictive policing technologies \jf{with a fraught history of social justice concerns around policing}.}  
\jf{An improved understanding of these questions could better inform policy decisions around predictive policing technologies in Baltimore, and beyond. Therefore, in this work, we perform a comprehensive comparative simulation study on the fairness and accuracy of predictive policing technologies in Baltimore.}  
\jf{Our results suggest that the situation around bias in predictive policing is more complex than was previously assumed. While  predictive policing exhibited bias due to feedback loops as was previously reported~\cite{lum2016predict,ensign2018runaway}, we found that the traditional alternative, hot spots policing, had similar issues. Predictive policing was found to be more fair and accurate than hot spots policing in the short term, although it amplified bias faster, suggesting the potential for worse long-run behavior. In Baltimore, in some cases the bias in these systems tended toward over-policing in White neighborhoods, unlike in previous studies~\cite{lum2016predict}.}
\jf{Overall, this} work demonstrates a 
\jf{methodology} for 
\jf{city-specific} evaluation and behavioral-tendency comparison of predictive policing systems, 
showing how 
\jf{such} simulations can reveal inequities and long-term tendencies. 
\jf{We recommend that authorities and community stakeholders use simulation methodologies to assist in collaboratively navigating the complexities around fairness in predictive policing.} 
\end{abstract}





\section{Introduction}
Advancements in AI have led to the development of predictive policing systems, with the 
\jf{potential} of high\jf{er} accuracy in predicting crime~\cite{zubair2025crime}. Predictive policing has been shown to have improvements over the simpler prior approach~\cite{mohler2011self}, called \textbf{hot spots policing}, also known as hotspot policing~\cite{braga2019hot, bowers2004prospective}, a strategy that \jf{uses data and simple statistical models to find} areas with high crime rates and allocates officers accordingly. This strategy is based on the assumption that crime distribution is affected by environmental factors, such as a neighborhood not being patrolled enough or one with profitable burglary targets, as well as the successful crime attempts of the criminals, so where crime was observed, there is more chance of it happening again~\cite{chapman2022data}. These simple models have evolved into more sophisticated \textbf{predictive policing models} that could take into account more factors, including the changes in crime distribution over time and predict future crime distribution~\cite{perry2013predictive}. These predictive models promise higher accuracy but whether they are fairer or not remains a\jf{n important} question. Previous research has suggested that predictive policing models are racially biased, but has not compared the extent of bias to traditional hot spots policing methods~\cite{lum2016predict}\jf{.} Therefore, in this work, our comparative study provides bases upon which these models' performances could be evaluated against one another.

Along with comparing the performance of predictive policing and hot spots policing, we investigated the long-term potential of using these methods in Baltimore City\jf{, Maryland, USA}.
Baltimore provides an interesting and important context for several reasons: 
\begin{itemize}
    \item Its history of reports of violence~\cite{ACLU2010Settlement, ACLU2012NonCompliance, DOJ2016Findings}, 
    \item The evident geographical racial divide around the city~\cite{brown2021black, pietila2010not}, 
    \item The animosity that arose among police and citizens around 2015, after the case of Freddie Carlos Gray Jr~\cite{Time2015FreddieGray, VanityFair2015FreddieGray}, and 
    \item Its law enforcement’s use of predictive policing models, starting around 2018~\cite{BaltimoreSun, FoxBaltimore}.
\end{itemize}
Neighborhoods in Baltimore remain largely segregated but not merely as an after-effect of historical segregation policies, which resulted in the \jf{``W}hite flight,\jf{''} \jf{k}eeping the Black majority in inner Baltimore and driving the White residents toward \jf{outer} suburbs~\cite{hanlon2007fate}. \jf{According to the ``New Jim Crow'' theory~\cite{alexander2012new}, t}he mass incarceration in Black neighborhoods functions as the modern way of maintaining segregation by ensuring  people of color reside in the low-income part of the city~\cite{alexander2012new}. This segregated status of neighborhoods reinforces racial and economic disparities among the neighborhoods, which in turn makes it 
\jf{increasingly vital to address} the fairness of predictive policing algorithms.

What complicates fairness evaluations further is the difference in the effect of police presence in Black neighborhoods vs. White neighborhoods. As Alexander, in her book ``The New Jim Crow,'' points out, policing in Black neighborhoods is militarized to the extent that Black youth started referring to police presence in their community as ``The Occupation\jf{.}'' Alexander mentions that the aggressive tactics used in Black neighborhoods are not even newsworthy, while the same tactic would be a ``political suicide'' in White urban neighborhoods. According to Alexander, the reason for this is the lower political influence of the Black community~\cite{alexander2012new}. 

These policing tactics affect the psychological and social atmosphere in these neighborhoods. In White communities, police presence is often perceived to bring safety and order. In Black neighborhoods, however, it causes fear, anxiety, and instability—conditions that can escalate tension and would end up with aggression, both by police forces and residents. Therefore, the same presence can have varied effects in different communities~\cite{alexander2012new}. Policing decisions made using predictive models' results might reinforce harm rather than reduce crime if they do not consider these dynamics. Pre-evaluation of these systems uncovers their police-concentration tendencies, which could help with better-informed policing decision-making.

The death of Freddie Carlos Gray Jr., a 25 year-old Black man, caused by the injuries sustained while in police custody, led to riots and city-wide protests, heightening the tension between the police and the Black community~\cite{DOJ2016Findings}. The protests were rooted in the over-policing of Black neighborhoods by heavy surveillance, wrongful arrests, and the under-protection with delayed and insufficient police response to calls for service~\cite{NoBoundaries2016Report, densley2021over}.
\jf{I}n the wake of these protests, there was a spike in crime rates, especially in homicides and non-fatal shootings, in Baltimore~\cite{cbs2015freddiegray, fivethirtyeight2015baltimore}. Two reasons have been reported as probable causes of this spike:
\begin{enumerate}
    \item The police pullback: the police department had a month-long retreat in response to the protests~\cite{cbs2015freddiegray}. There was a drop in the number of arrests~\cite{fivethirtyeight2015baltimore}.
    \item Opportunists and emboldened criminals: A public survey by the PEW Research Center showed that 60\% believed the main cause of the crime rise was that some people took advantage of the situation~\cite{pew2015baltimore}.During the unrest and subsequent police pullback, opportunists looted approximately 30 pharmacies and drug clinics, stealing substantial amounts of narcotics, which some believed caused a rise in crimes as well~\cite{cbc2015baltimore,latimes2015baltimore, cnn2015baltimore}. 
\end{enumerate}
The grassroots movements following the death of Freddie Gray and the mass arrests of the protesters~\cite{Guardian2016FreddieGrayTimeline} resulted in the establishment of several community-led organizations advocating for police reform and racial justice, namely, the Baltimore Action Legal Team (BALT)~\cite{BALT},  the No Boundaries Coalition Community~\cite{NoBoundaries}, and Leaders of a Beautiful Struggle (LBS)~\cite{LBS}. These local organizations, with the help of activists, lawyers, and community members, are trying for reform by providing legal assistance, public education, and policy advocacy for Black residents affected by biased policing practices. Following the grassroots movements' goals, our study extends their vision by addressing the technological aspect of policing reform.

The segregated neighborhoods of Baltimore, along with its history of aggressive law enforcement tactics, could have taken root in Baltimore's crime records, on which predictive policing algorithms get trained. Baltimore's law enforcement has been using predictive policing tools without prior evaluations and analysis, estimating potential long-term performance tendencies in Baltimore's context, which is a problem that needs to be addressed.

With this study, our contributions, therefore, are\jf{:}
\begin{enumerate}
    \item A simulation study of the fairness of predictive policing in the context of Baltimore,
    \item A comparative analysis of hot spots policing and predictive policing,
    \item Introducing a logical basis on which predictive policing models could be compared, 
    \item A visualization of results with suitable graphics assisting the comprehensibility of the comparisons and the interpretability of related findings, 
    \item Last and most importantly, demonstrating the necessity and providing a method for localized evaluation of policing systems prior to real-world implementation.
\end{enumerate}
In our 300 day-simulation period, predictive policing develops less bias on average compared to hot spots policing while maintaining higher accuracy, but it could change if the simulation is run for longer because of \jf{p}redictive policing's higher speed of bias development. Contrary to the prevailing wisdom, in Baltimore the feedback loop trends toward assigning higher average officer share to White neighborhoods for Kernel Density Estimation (KDE) and PredPol when trained on all crime records from 2018-2019.

\section{Related Work}

Since the emergence of predictive policing systems, research in this field has focused on four main areas: accuracy improvement~\cite{mohler2011self, mohler2014marked, hu2018spatio}, fairness enhancements~\cite{lum2016predict, ensign2018runaway, mohler2018penalized}, comparative studies~\cite{mashiat2023counterfactually, griffard2019bias}, and contextualized implementation studies~\cite{lum2016predict, mohler2018penalized, repasky2024multi, barbosa2015exploiting, akpinar2021effect}, with some works contributing to more than one aspect.

\subsection{Accuracy Improvement}
Advancements in predictive policing have grown with improved data collection methods and more sophisticated techniques. Earlier predictive systems in this field relied on hotspot mapping methods such as Kernel Density Estimation (KDE)~\cite{parzen1962estimation, Rosenblatt1956RemarksOS} to predict crime densities~\cite{chainey2008utility}. The temporal-aware crime mapping claimed accuracy improvement over the non-aware static ones by taking into account the time distance of prior crimes to the prediction target time~\cite{bowers2004prospective}. The spatio-temporal KDE, by accounting for the time and space dependency of crime distribution, was proposed to improve hot spots policing, setting the foundation for more enhanced predictive models~\cite{hu2018spatio}.The epidemic-type aftershock sequence (ETAS) crime forecasting model, making use of the expectation maximization (EM) algorithm, improved the accuracy of assigning crime risk scores to different geographic areas~\cite{mohler2011self, mohler2014marked}. This ETAS-based framework was adopted and operationalized by the PredPol predictive policing system to generate hotspot forecasts~\cite{uspatent8949164}.

In more recent years, we have seen the use of textual data for crime analysis and prediction using data from social media~\cite{vivek2023spatio, tam2023multimodal}. Vivek and Prathap~\cite{vivek2023spatio} used crime-related tweet counts and compared the predictions of the crime counts by LSTM, ARIMA, and SARIMA. They showed that ARIMA performed best and LSTM performed worst at predicting crime counts, probably due to limited training data. They showed that tweet crime counts fairly matched real-world crime counts and that they could be used as a proxy for real-world crime data in certain crime categories. S. Tam et al.~\cite{tam2023multimodal} combined twitter sentiment analysis with historical crime data and proposed the ConvBiLSTM model. The ConvBiLSTM model integrates CNN and BiLSTM networks. They fused the processed textual and non-textual data and fed it into a Bidirectional LSTM. It predicted whether a tweet could be indicative of a crime with about 97\% accuracy.

Reinforcement learning has contributed to the enhancement of police allocation logics~\cite{barbosa2015exploiting, joe2022reinforcement, chen2023risk, chen2023optimizing, repasky2024multi, joe2023learning}. Barbosa et al. proposed a model-free RL approach to optimize patrol agent position. The model learns from digital pheromones left from previous events. After repositioning an agent, it gets rewarded if an agent is present where a crime is happening and penalized if a crime happens where there is no patrol~\cite{barbosa2015exploiting}. Joe et al.~\cite{joe2022reinforcement} use deep reinforcement learning and temporal-difference learning for patrol and incident response. Their model learns dispatch and rescheduling jointly with two main objectives: maximizing successful incident response and minimizing patrol schedule disruption. Another work on patrolling improvement uses IDQN and deep reinforcement learning to learn dynamic patrol routes by maximizing patrol coverage and prioritizing high-risk areas. The city was represented by a graph, with locations as nodes and roads as edges. Each road has two attributes: crime risk and distance. Patrolling a high-risk road was set to be rewarded, while revisiting a road diminished the reward~\cite{chen2023risk,chen2023optimizing}. Multi-agent reinforcement learning (MARL) was also used to optimize patrolling and dispatching by a joint policy. The model is rewarded if the decision gives a faster response and better coverage and is penalized if it results in delays or gaps in patrols~\cite{repasky2024multi}.

The aforementioned studies do not address fairness; however, other studies evaluate fairness, as we discuss in the subsection below.

\subsection{Fairness Studies}
In 2016, the unfairness tendencies of these predictive policing models were brought to light when Lum and Isaac’s simulation study~\cite{lum2016predict} on Oakland’s drug-related crime demonstrated how police would frequent minority and low-income neighborhoods when these models were applied. This motivated more studies focusing on fairness analysis of these systems. 

Brantingham ran a study, explaining how the bias gets into the data when an officer downgrades a crime or upgrades a crime based on personal biases~\cite{brantingham2017logic}. In this study, simulations were run once downgrading crime counts by removing a percentage of crimes from the actual dataset and once upgrading by adding a percentage more of crimes. Five increasing percentage numbers were used (2, 5, 10, 15, and 20 percent). They compared model parameters and demonstrated that for downgrading, the bias should be substantial to have a noticeable impact on the model parameters. By \jf{``}noticeable impact,\jf{''} they meant the parameters' values estimated more than one standard deviation away from what they 
\jf{would} have been if there was no downgrading. But for upgrading, the noticeable impact happened on a lower upgrading percentage (10 percent)~\cite{brantingham2017logic}. In 2018, Ensign et al. 
\jf{mathematically analyzed how} this bias could happen using a Polya-urn model in 4 different scenarios of allocating one officer to one of the two neighborhoods:  
\begin{enumerate}
    \item High-crime-rate neighborhood vs. low-crime-rate training over only discovered crimes, 
    \item The same this time training overreported crimes. 
    \item Two high-crime-rate neighborhoods training over only discovered crimes, 
    \item The same neighborhoods being trained on reported crimes~\cite{ensign2018runaway}.
\end{enumerate}
They proposed 
\jf{several} remedies that could improve that specific model~\cite{ensign2018runaway}. The same year, Brantingham, Mohler, et al. 
\jf{investigated} evidence of bias in their randomized controlled trial of the ETAS model, which 
\jf{their results suggested was}  nonexistent~\cite{brantingham2018does}. However, their bias measure is different from what is measured in the fairness of police distribution when using predictive policing algorithms in simulations~\cite{brantingham2018does, akpinar2021effect, chapman2022data}. They didn't look at police concentration after long-term use of the predictive policing model and the effect of the feedback loop. Their bias detection focused on detecting significant racial differences in arrests between control (days 20 hotspots were chosen by the human analysts) and treatment days (days 20 hotspots were chosen by Predictive Policing algorithm)~\cite{brantingham2018does}.  However, Mohler introduced a new version of the ETAS model in another paper, one with a penalized likelihood method that incorporated demographic parity and demonstrated the give-and-take between accuracy and fairness~\cite{mohler2018penalized}. 

Akpinar et al. contributed to the fairness discussion by demonstrating that even by using victim-reported data instead of arrest data (i.e., discovered crime), the models would show bias because in some areas crimes are under-reported compared to others~\cite{akpinar2021effect}. They pointed out that the unfair rankings were \textbf{data-driven} and not model-specific. They used a synthetic crime dataset generated by the Self-Exciting Point Process model (SEPP) over Bogota. They thinned the generated dataset by the victimization rate of each district in Bogota to form the synthetic true crime dataset and then thinned it further using the victim report rate of each district to create the reported crime dataset. A simulation was run once using the Moving Average (MAVG) model and once using the SEPP model. It was shown that in both cases, fewer hotspots were being predicted in low-reporting districts, regardless of the actual crime rate. While this study demonstrates the inherent risk of data-driven bias even across different model types, it does not simulate police deployment or detection feedback. Nor does it compare models in terms of predictive performance. Our work builds on this insight by using real-world crime data, modeling officer deployment and detection, and examining how retraining based on observed and reported crime influences the fairness and accuracy of hot spots policing and predictive policing over time~\cite{akpinar2021effect}.

Chapman et al., in 2022, attempted to argue that the cause of bias is the theoretical assumptions behind model development~\cite{chapman2022data}. They 
\jf{assert that} criminological theories are categorized in two beliefs:
\begin{enumerate}
    \item The \textit{neo-classical}: crime is the product of rational choice and police presence is preventive. One successful burglary could motivate another one somewhere close by so there is a relation among crimes within space and time.
    \item The \textit{positivists}: crime is based in genetics, social setting, and biology and not by choice, so its distribution is mostly random.
\end{enumerate}
To show that this bias arises from the model. They create three sets of synthetic crime data: one with a uniform and random distribution, the second mostly uniform with two hotspot regions added, and the third real-world data from Kent. They attempt to show that simulating \jf{P}redpol over each of these datasets caused clearer hotspots to form after 48 days of simulation. Police were distributed among \jf{the} 10 top-ranked regions. 

\jf{We note} that even if the model had ranked them the same, only 10 regions would 
\jf{be allocated} police. From then on, hotspots appear. 
\jf{It is unclear whether this demonstrates that the bias was caused by the model, or by limited police resources.} 
They also tried removing the effect of feedback in a simulation using the random crime dataset. They observed that no hotspots were formed, countering their 
assumption that the model was the cause of bias. 
\jf{In this experiment, they changed the} data that is fed into the model and not the model itself. They conclude that bias is \textbf{model-driven}, unlike Akpinar~\cite{chapman2022data, akpinar2021effect}. In 2023, Mashiat et al., in an extended abstract, proposed the use of a counterfactual causal model to assure fairness. They compared their model, PropFair, with Ensign’s PolyaUrn model by police allocation between two neighborhoods with constant crime rates of 0.3 and 0.7 and claimed theirs matched the fairness line better, allocating about 4.5 fractions of police (out of 10 police) to beat 1 and the rest to beat 2~\cite{mashiat2023counterfactually}. There have also been fairness studies on some investigative tools~\cite{griffard2019bias} and also criminal recidivism prediction systems~\cite{lagioia2023algorithmic, wang2019empirical, dressel2018accuracy}. Here we 
\jf{focused on} fairness studies on crime prediction and patrolling-related tools.

\subsection{Comparative Analysis}
There are numerous works comparing the accuracy of two models with noticeable differences in their underlying logic~\cite{chen2023risk, joe2022reinforcement, tam2023multimodal, bowers2004prospective, hu2018spatio, joe2023learning, barbosa2015exploiting}. Among which we can see those proposing new models for incident response or route-patrolling scheduling~\cite{chen2023risk, joe2022reinforcement, barbosa2015exploiting, joe2023learning} and some others proposing crime-prediction models helping with policing decisions~\cite{tam2023multimodal, bowers2004prospective, hu2018spatio, mohler2011self}. To show the enhancement in the accuracy by their proposed framework, these studies compare their proposed model to other models or to the same base model not optimized considering a specific feature.

Others focus on comparing the unbiased version of an algorithm with its original version in terms of fairness~\cite{repasky2024multi, mohler2018penalized, ensign2018runaway}. Ensign theoretically compared Predpol with a modified version of it in allocating an officer between two regions~\cite{ensign2018runaway}. Mohler~\cite{mohler2018penalized} compared a Hawkes process-based model to a version of it with a penalized likelihood, while Repaskey studied fairness in the officer-response time of a Multi-Agent Reinforcement Learning model (MARL) and an optimized one by rewarding even patrol coverage~\cite{repasky2024multi}.

There are some existing works that compare both the fairness and accuracy of two different algorithms with different levels of complexity.
Mashiat, in their extended abstract, did a theoretical fairness and accuracy comparison between \jf{P}ropFair, their proposed counterfactual causal model, and Ensign’s Polya Urn model, using the same two-region police allocation scenario as Ensign’s 2018 study~\cite{mashiat2023counterfactually}. Griffard et al. studied fairness and accuracy in finding a pattern between a newly police-reported crime and previous cases by looking at the rate at which the patterns were found in minority neighborhoods versus others. They compared Patternizr, an investigative tool deployed by the NYPD, with some baseline models such as Gradient boosting decision trees~\cite{griffard2019bias}.

\subsection{Contextualized Studies}
There have also been a number of studies analyzing the performance and impact of predictive policing systems in a certain region, some through simulation and some through real-world experiments. The simulation studies have investigated models' behavior in cities like Oakland, Indianapolis, Atlanta, Denver, and Bogota~\cite{lum2016predict, mohler2018penalized, repasky2024multi, barbosa2015exploiting, akpinar2021effect}. The real-world implemented performance analysis was in LA (USA), Kent (UK), and Memphis~\cite{brantingham2018does, mohler2015randomized, helms2020memphis}. Baltimore has not been systematically studied before, although the city is making use of AI tools in their law enforcement now more than before. We believe it is imperative to be aware of the unfairness tendencies of such models specific to the context before real-world application, since any city policies can further impact the demographic distribution and be a cause of economic, social, and racial divide as seen in the past~\cite{hanlon2007fate}. Therefore, one of the aims of this study is to analyze model performance in Baltimore, an important deployed real-world context which is also representative of areas with a history of racial segregation.

As these studies show, fairness distortions in predictive policing systems can emerge from both modeling assumptions and biased data inputs. Yet few have examined how these biases evolve over time in real-world deployment settings, another gap we aim to address through simulation in the Baltimore context. For more information on the body of work in this field, see previous literature reviews by Raji et al.~\cite{rajipredictive} and Mandalapu et al.~\cite{mandalapu2023crime}.

\section{Materials and Methods}
In this section, we describe the datasets used and some basic terminology that would facilitate the explanation of the simulated scenarios, followed by the details of the simulations.

\subsection{Data Description}
We use the crime and neighborhood dataset available on the Open Baltimore website~\cite{baltimore_crime_data, baltimore_neighborhood_data, baltimore_neighborhood_boundaries}.\footnote{Data Source: \url{https://data.baltimorecity.gov/datasets/}} As the Open Baltimore website periodically updates these datasets and its directory structure may change, we also maintain archived copies to ensure reproducibility.\footnote{Downloaded version: \\
\url{https://github.com/saminsemsar/Data_Analysis_Portfolio/tree/main/PredictivePolicing/Data}} 
The crime dataset includes information such as the date and time of the crime, latitude, longitude, neighborhood, and crime description. For the study, crime locations outside residential areas were excluded. After performing data preprocessing, a total of 207,447 crime records that occurred in Baltimore City in the years 2019-2022 were utilized. Note that these crime records are 
only a subset of the incidents that occurred in that period, as they are the only ones documented and known to law enforcement, \jf{and this is likely to impact the behavior and potential bias in predictive policing algorithms~\cite{lum2016predict}}.

\subsection{Definitions}\label{subsec: Definitions}
In this paper, several key terms will be \jf{used}, 
 which we have defined below.
\begin{itemize}
    \item \textbf{Black neighborhoods:}\label{def: black-neis} We define a neighborhood as \textit{Black} if the number of Black residents in that neighborhood exceeds the number of residents of any other racial group. For example, if a neighborhood has 21 Black residents, 20 White residents, and 10 Latino residents, it is flagged as a Black neighborhood. This majority-based definition allows us to consistently categorize neighborhoods by their predominant racial composition for fairness analysis.
    \item \textbf{White neighborhoods:}\label{def:white-neis} Neighborhoods with  majority \textit{White} residents.
    \item \textbf{Neither-Black-Nor-White neighborhoods:}\label{def:other-neis} Neighborhoods where the majority of their residents are neither Black nor White. These areas are populated primarily by groups including the \jf{Latinx community,} 
    Alaskans, and other racial or ethnic minorities.
    
    \item \textbf{Noisy-OR:}\label{def:noisy-or}
    We use a Noisy-OR function to model the probability of an effect \( Y \) given binary causes \( X_1, \ldots, X_n \). \( X_i\) indicates whether the cause \(i\) is present or not. Each cause \( X_i \) is associated with a failure probability \( q_i \). The Noisy-OR formulation assumes conditional independence and is defined as:
    \begin{equation}
    P(Y = 1 \mid X_1, X_2, \ldots, X_n) = 1 - \prod_{i=1}^{n} q_i^{X_i} \mbox{ .}
    \label{eq:noisy-or}
    \end{equation}

   We incorporate the Noisy-OR, a probabilistic model, to introduce uncertainty into the crime detection. The Noisy-OR model determines an event (e.g., crime detection) to have occurred if any of several coin flips come up as \emph{heads}. If a coin flips ``heads,'' it is marked as ``detected'' based on the context. We use Noisy-OR to add some 
   uncertainty to the \jf{crime detection} probabilities\jf{, reflecting the notion that if any of the officers in the vicinity detect a crime, then it is detected}. 
   
    \item \textbf{Detected crimes:}\label{def:detected-crimes} The algorithm determines whether a crime is detected by applying the Noisy-OR model to the probability of detection (p), taking into account the number of police officers within the neighborhood $k$ and the probability of an officer detecting a crime, which is a hyperparameter $p = 0.5$. 
    \begin{equation}
        P(\text{Crime Detected}) = 1 - (1 - p)^{k} 
        \label{eq:noisy-or-detected}
    \end{equation}
    Then the crime is labeled as detected based on a coin flip with the probability in~\cref{eq:noisy-or-detected}.

    \item \textbf{Reported crimes:}\label{def:reported-crimes} We created a report dataset by flipping a coin with 0.4 probability to decide if each crime would be reported or not~\cref{eq:coin-flip-Reported}. The number 0.4 was calculated using a weighted average on the number of each type of crime in our dataset and the probability of reporting that crime, based on the Bureau of Justice Statistics (BJS), 2019 report~\cite{bjs_crime_2019}. We weight each crime category’s count by its national reporting probability from the BJS. The estimate is computed as:

    \begin{equation}
        \text{Average Report Probability of Crime} = \frac{\sum\limits_{i \in \mathcal{CT}} N_i \cdot r_i}{\sum\limits_{i \in \mathcal{CT}} N_i}\mbox{ .}
        \label{eq:crime_prob}
    \end{equation}

    \begin{equation}
        \text{Crime is Reported if } (0.4> \text{ rand})
        \label{eq:coin-flip-Reported}
    \end{equation}
    
    Here, $N_i$  is the number of crimes of the type $i$ that occurred after 2019 in Baltimore, and $r_i$ is the national reporting rate of the crime type $i$ (see \cref{fig:BJS}).
    \begin{figure}[H]
        \centering
        \includegraphics[width=\textwidth]{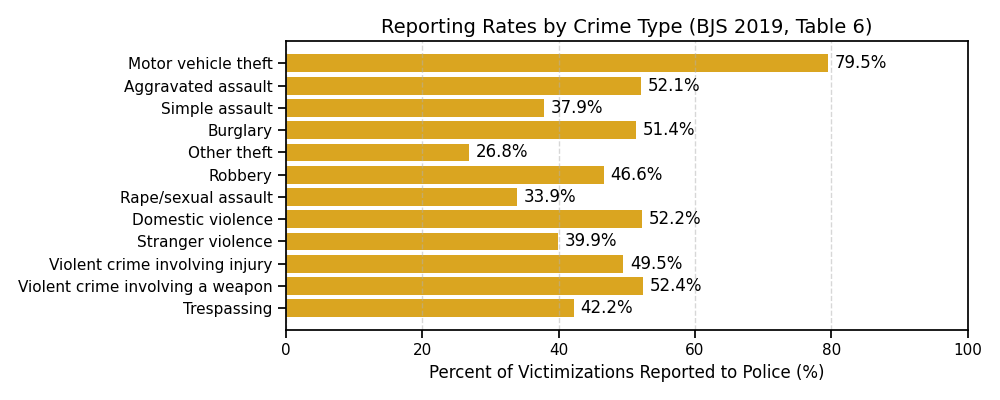}
        \caption{Crime-report probabilities based on data from Bureau of Justice Statistics, 2019 report~\cite{bjs_crime_2019}}.
        \label{fig:BJS}
    \end{figure}
    \item \textbf{KDE:} 
    Kernel Density Estimation \jf{(KDE)} is a statistical technique used to estimate the underlying Probability Density Function (PDF) of a set of data points. It involves creating a smooth continuous function by placing a kernel (a predefined shape, such as \jf{a} Gaussian) on each data point and summing them up. The resulting estimated density function provides insights into the distribution and intensity of the data across the entire range. We implemented Kernel Density Estimation 
    using the KernelDensity class from the scikit-learn Python library \cite{scikit_learn}, which provides a computationally efficient implementation of the density estimation method originally introduced by Rosenblatt \cite{Rosenblatt1956RemarksOS} and Parzen \cite{parzen1962estimation}.
    \item \textbf{Short-term KDE:}\label{def:short-kde} KDE which receives short-term crime history. In our simulation we have set this short-term to equal a month's worth of crime history.
    \item \textbf{Long-term KDE:}\label{def:long-kde} KDE which receives a longer history of crime compared to short-term KDE. In our simulations we have set it to be equal to one year's worth of crime history to make it comparable to PredPol in terms of the data they receive.
    \item \textbf{PredPol:}\label{def:pred} PredPol is a self-exciting point process model for crime prediction introduced by Mohler et al. \cite{mohler2011self,mohler2015randomized}. The model is based on the Epidemic-Type Aftershock Sequence (ETAS) framework originally developed in seismology, in which each past event increases the short-term likelihood of nearby future events. In the context of crime forecasting, this formulation captures near-repeat victimization by modeling crime intensity as a combination of a background rate and a self-exciting component. We implemented the simplified formulation explained in Mohler's 2015 work~\cite{mohler2015randomized}.

\end{itemize}

\subsection{Methods}
\label{sec:methods}
A simulation-based framework was developed to investigate how policing algorithms perform 
\jf{over time} with respect to fairness and accuracy. The real crime data and neighborhood data from Baltimore were applied in the simulation, which includes neighborhood identifiers, timestamps, and racial demographic indicators. Our goal is to observe and quantify the temporal dynamics of bias and accuracy in policing models. For this purpose, during the course of the simulation, 3 variables were calculated in each scenario setting, for each day, and for each neighborhood:
\begin{enumerate}
    \item number of real crimes in Baltimore city dataset,
    \item number of police assigned,
    \item number of crimes detected.
\end{enumerate}
The simulation was run 20 times in each scenario setting. Therefore, the number of police assigned and the number of crimes detected were averaged over 20 runs. 

Since the number of ``Neither-White-Nor-Black'' neighborhoods was very limited in Baltimore (only three), they were discarded from the analysis to enable meaningful comparative analysis. 
The simulations involved running PredPol, short-term KDE, and long-term KDE over \textbf{eight} distinct settings. PredPol is a representative of predictive policing models, while short-term KDE and long-term KDE are representatives of hot spots policing. PredPol is applied to a dataset of all known crimes to predict future crime distributions. This model gives higher weights to more recent crimes. \jf{H}ot spots policing, on the other hand, typically receives only recent crimes and only calculates the crime density of different areas. This method does not perform any time-wise calculation. 
\jf{Short-term KDE thus more accurately represents the practice of hot spots policing than long-term KDE. However, we include long-term KDE for a fair comparison with PredPol, as it uses the same amount of data, while short-term KDE only uses the most recent data.} 
These algorithms were simulated over various settings defined by three variables:
\begin{enumerate}
    \item Number of police officers: distributing 40 or 400 officers,
    \item Probability of reporting a crime: using only detected crimes (probability of report = 0) or detected and reported crimes (probability of report = 0.4),
    \item Crime type: using all types of crime in the dataset ('TOTAL') or only aggravated assault records ('AGG. ASSAULT').
\end{enumerate}

PredPol and long-term KDE used all available detected and reported crime data, while short-term KDE used detected and reported crimes within one month before the prediction date.

During the simulation\jf{,} police locations were determined for each day starting from January 1, 2019, for 300 days. 
Note that the police assignment for the first date was determined by considering all the real crimes that occurred before that date, assuming that all the crimes were detected. However, for the subsequent days, we appended that crime history to a subset of the real crimes from the previous days that were specifically labeled as detected by the simulation, plus the reported crimes of the previous days. 
Noisy-OR was applied to the number of police officers within the neighborhood in which each crime had happened to generate the detected crimes dataset for the current day. This dataset, along with the reported crimes dataset, was appended and then used to determine police locations for the following day. The pseudo-codes in~\cref{alg:ReportedCrimeDataset} and~\cref{alg:simulation} clarify our strategy.

\begin{algorithm}[t]
    \caption{Creating a Reported Crimes Dataset}
    \textbf{Input:} Crime dataset $crimes$, Report probability $report\_probability$ \\
    \textbf{Output:} None\\
    $reported\_crimes$ = an empty crime dataset\;
    \ForEach{crime $c$ in $crimes$}{
        flip a coin with a report probability\;
        \If{heads : $random > report\_probability$}{
            $reported\_crimes = reported\_crimes + c$
        }
    }
    save CSV file of $reported\_crimes$;
    \label{alg:ReportedCrimeDataset}
\end{algorithm}

\begin{algorithm}[t]
    \caption{Police Allocation Simulation}
    \KwIn{Crime dataset $crimes$; prediction algorithm $Alg$; start date $start\_date$; end date $end\_date$; detection probability $detectProb$; reported-crimes dataset $reported\_crimes$}
    \KwOut{Result dataset $res$ containing daily neighborhood-level police allocations and detected crimes}
    
    Initialize $crimes\_h \gets$ all crimes before $start\_date$\;
    Initialize $detected\_crimes \gets crimes\_h$\;
    Initialize $res$ with columns: \emph{date}, \emph{neighborhood}, \emph{crime\_num}, \emph{police\_num}, \emph{detected\_crime\_num}\;
    Fill $res$ with simulation dates, neighborhoods, and number of crimes from $crimes$\;
    
    \For{$d \gets start\_date$ \KwTo $end\_date$}{
        \uIf{$Alg =$ short-term KDE}{
            $h \gets$ recent 1-month crime history\;
        }
        \Else{
            $h \gets$ all history\;
        }
        $crimes\_h \gets reported\_crimes\_h \cup detected\_crimes\_h$\;
        Remove duplicates from $crimes\_h$\;
        $crime\_Model \gets$ fit $Alg$ to $crimes\_h$\;
        Sample police locations $police\_locations\_d$ from $crime\_Model$\;
        Update $res$ with police numbers per neighborhood on date $d$\;
    
        \ForEach{crime $c \in crimes\_d$}{
            $numPolice \gets$ number of officers in $c$’s neighborhood\;
            $p(detected) \gets 1 - (1 - detectProb)^{numPolice}$ \tcc*[r]{noisy-OR}
            \If{random $< p(detected)$}{
                $detected\_crimes\_d \gets detected\_crimes\_d \cup \{c\}$\;
            }
        }
        Update $res$ with detected crime counts for each neighborhood on date $d$\;
    }
    \Return{$res$}\;
    \label{alg:simulation}
\end{algorithm}

\subsection{Fairness and Accuracy Metrics}
The gathered results from running KDE and PredPol in different scenarios were analyzed using different visualizations and statistical assessments. 
\\To improve legibility and reduce redundancy, we use both the notation $\overline{x}$ and $Avg(x)$ to denote the arithmetic mean (i.e., the sum of values divided by the number of elements). These notations are used interchangeably throughout the formulas, depending on which improves clarity and visual flow.

In our data analysis to quantify fairness and accuracy, we focused on five metrics to analyze three major concepts: the racial fairness gap, the neighborhood-level fairness gap, and coverage accuracy. For both \jf{r}acial group fairness and individual neighborhood fairness, equality of treatment is a combination of having equal resources in general  and equal resources given to similar individuals. For that purpose, in both racial and neighborhood-level fairness, we 
\jf{consider} the equality of police share or police officer number as general resource equity and the equality of police share to crime share ratio as a measure for equity in treatment of individuals having similar crime rates, i.e., treatment proportional to crime rate. 
\paragraph{The most fair model or simulation setting in terms of any fairness-gap metric is defined as the one with the minimum average value for that metric during simulation, while the most accurate one is defined as the one with the maximum average accuracy value during the simulation.}\label{def:most-fair-or-accurate}
\begin{enumerate}
    \item \textbf{Racial Fairness Gap}\\
    Racial fairness in this study is a group fairness \jf{metric}. \jf{In this work,} racial disparity or the fairness-gap 
    is defined as 
    \jf{an} absolute difference in averages. \jf{The use of a} difference or absolute difference is a standard approach in quantifying group disparity (
    \jf{cf.} Equation~5 in~\cite{dwork2012fairness} and Definition~6.3 in~\cite{friedler2019comparative}).
    \begin{itemize}
        \item \textbf{Inequality of Average Police Share between White and Black Neighborhoods}\\
        This measure determines the disparity of average police share between the two races, meaning the groups' general equality of resource allocation: 
        
        \begin{equation}
        \text{RacialFairnessGap}_{\text{PoliceShare}} = \left| \overline{P_{\text{Black}}} - \overline{P_{\text{White}}} \right|
        \label{}
        \end{equation}
        where \( \overline{P_{\text{Race}}} \) is the average police share in neighborhoods of the given race. Based on the explanation above, the most fair
        model or simulation setting would be determined by:
    
    
        \begin{equation}
        \text{MostFair}_{\text{PoliceShare}} = \arg\min \left(  Avg \left( \text{RacialFairnessGap}_{\text{PoliceShare}} \right) \right)
        \label{}
        \end{equation}

        \item  \textbf{Inequality of Average Police-to-Crime Ratio (PCR) between White and Black Neighborhoods}\\
        Racial fairness gap is defined as the absolute difference of the average police-to-crime ratio between groups, analyzing proportional treatment or similar treatment of individuals with similar crime rates (see~\cref{Ineq-PCR1,Ineq-PCR2,Ineq-PCR3,Ineq-PCR4,Ineq-PCR5}, 
        \jf{Here, $N = |\text{Neighborhoods}|$).} 
    
        \begin{equation}
        PoliceShareSmoothed_i = \frac{PoliceNum_i +\varepsilon}{TotalPoliceNum + N\varepsilon} \quad,  \quad i \in Neighborhoods
        \label{Ineq-PCR1}
        \end{equation}
        \begin{equation}
        CrimeShareSmoothed_i = \frac{CrimesNum_i +\varepsilon}{TotalCrimesNum + N\varepsilon} \quad, \quad i \in Neighborhoods
        \label{Ineq-PCR2}
        \end{equation}
        \begin{equation}
        \text{PCR}_{\text{Race}} = Avg \left( \frac{PoliceShareSmoothed_i}{CrimeShareSmoothed_i} \right) \quad, \quad i \in Neighborhoods_{Race}
        \label{Ineq-PCR3}
        \end{equation}
    
        \begin{equation}
        \text{RacialFairnessGap}_{\text{PCR}} = \left| \text{PCR}_{\text{Black}} - \text{PCR}_{\text{White}} \right|
        \label{Ineq-PCR4}
        \end{equation}
    
        \begin{equation}
        \text{MostFair}_{\text{PCR}} = \arg\min\left( Avg \left( \left| \text{RacialFairnessGap}_{\text{PCR}} \right| \right) \right)
        \label{Ineq-PCR5}
        \end{equation}
    \end{itemize}

    \item \textbf{Neighborhood-Level Fairness Gap}\\
    To study inequality of treatment among individual neighborhoods, we \jf{use} 
 the Gini coefficient. This measure of inequality has been used in health~\cite{abeles2020gini}, education~\cite{thomas2003measuring}, and 
 \jf{especially the} economic-related literature~\cite{de2007income}. It measures the distance from the equality line. Therefore, the higher the Gini, the more unequal the values are.
    The Gini coefficient is calculated using the trapezoidal method. \\
    \begin{equation}
    Gini(X) = \frac{1}{n} \left(n + 1 - 2 \cdot \frac{\sum_{i=1}^n \sum_{j=1}^i x_{(j)}}{\sum_{i=1}^n x_{(i)}}\right) \quad, 
    \text{where $x_{(i)}$ are sorted values.}
    \label{}
    \end{equation}
    \begin{itemize}
        \item \textbf{Inequality of Police Distribution}\\
        This metric calculates the Gini coefficient of police numbers of each neighborhood during the simulation to see the overall inequality of the resource distribution.
        \begin{equation}
        \text{Gini}_{\text{Police}} = \text{Gini}(\{ P_i \}_{i=1}^{N})
        \label{}
        \end{equation}
    
        \begin{equation}
        \text{MostFair}_{\text{Gini,Police}} = \arg\min\left( Avg \left(\text{Gini}_{\text{Police}} \right) \right)
        \label{}
        \end{equation}
        
        \item \textbf{Inequality of Police-to-Crime Ratio (PCR) Across Neighborhoods}\\
        \jf{We measured t}he Gini coefficient of neighborhoods' police-to-crime ratio, 
        which is the inequality of the police distribution of similar individuals with similar crime ratios and the inequality of officer shares proportional to crime shares.
        
        \begin{equation}
        \text{PCR}_i = \frac{P_i + \varepsilon}{C_i + \varepsilon}, \quad
        \text{Gini}_{\text{PCR}} = \text{Gini}(\{ \text{PCR}_i \}_{i=1}^{N})
        \label{}
        \end{equation}

        \begin{equation}
        \text{MostFair}_{\text{Gini,PCR}} = \arg\min\left( Avg \left( \text{Gini}_{\text{PCR}}\right) \right)
        \label{}
        \end{equation}

    \end{itemize}

    \item \textbf{Coverage Accuracy — Proportion of Crimes Detected}\\
    Coverage accuracy 
    \jf{measures} the effectiveness of 
    \jf{the police distribution relative to the actual crime distribution.}
    It is defined as the \jf{proportion of detected crimes.} 

    \begin{equation}
    \text{CoverageAccuracy} = \frac{\text{Total Detected Crimes}}{\text{Total Crimes}}
    \label{}
    \end{equation}

    \begin{equation}
    \text{MostAccurate} = \arg\max\left( Avg \left( \text{CoverageAccuracy}\right) \right)
    \label{}
    \end{equation}

\end{enumerate}

\section{Results}
\label{sec:Results}

\begin{figure}[H]
  \centering

  \begin{minipage}{0.32\textwidth}\centering
    \textbf{PredPol}\\[0.3em]
    \includegraphics[width=\linewidth]{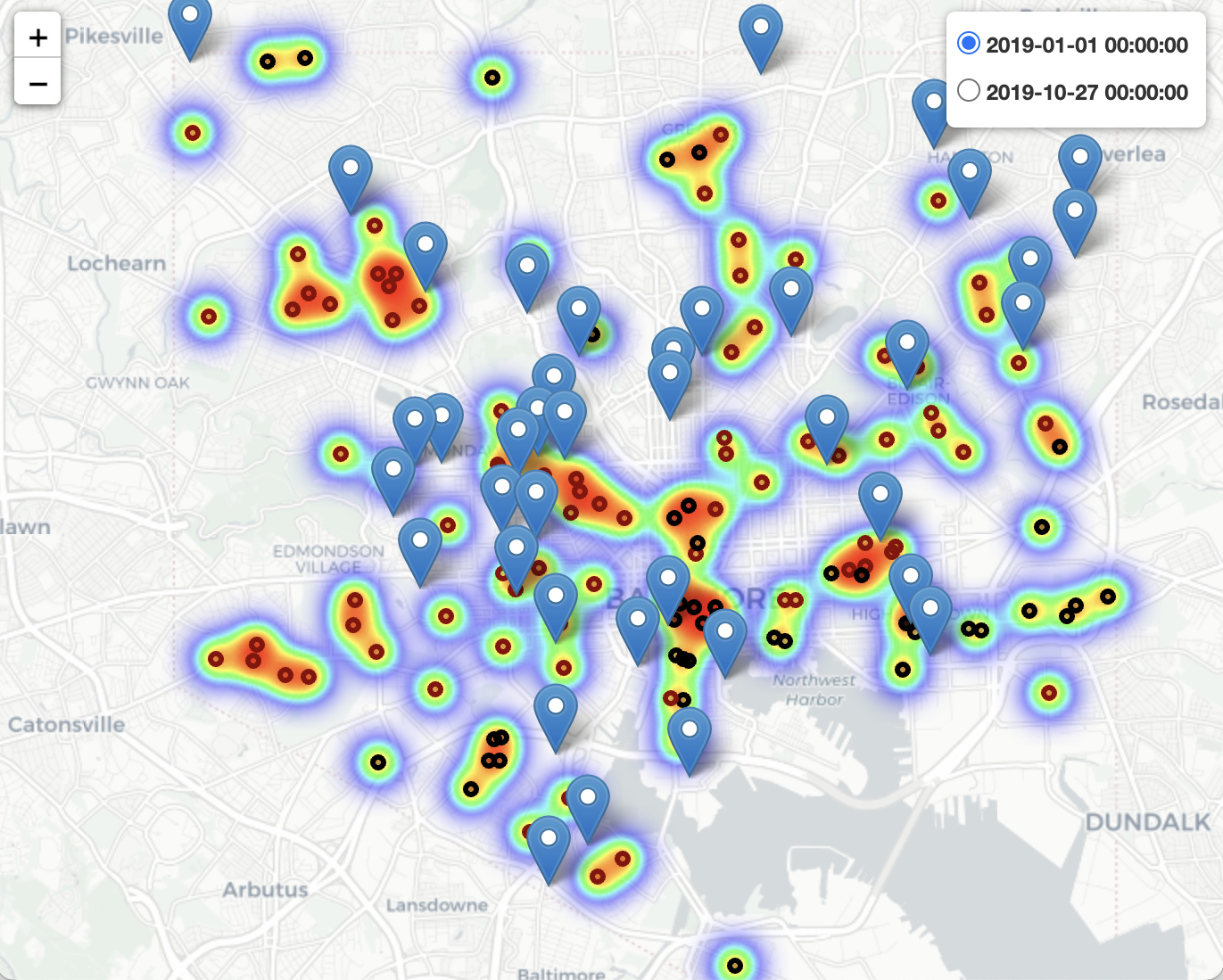}\\[-0.2em]
    \small Day 1
  \end{minipage}
  \hfill
  \begin{minipage}{0.32\textwidth}\centering
    \textbf{Long-Term KDE}\\[0.3em]
    \includegraphics[width=\linewidth]{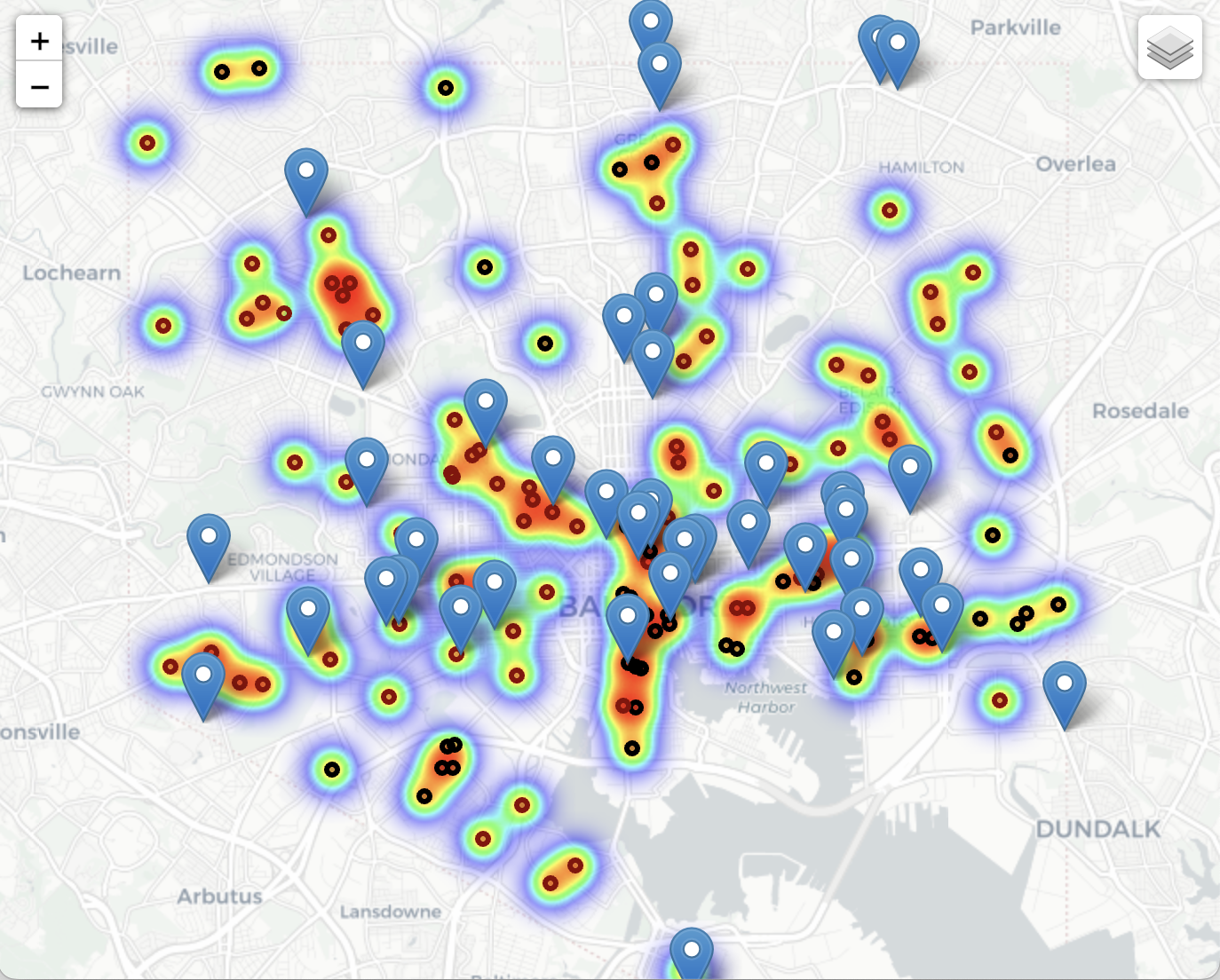}\\[-0.2em]
    \small Day 1
  \end{minipage}
  \hfill
  \begin{minipage}{0.32\textwidth}\centering
    \textbf{Short-Term KDE}\\[0.3em]
    \includegraphics[width=\linewidth]{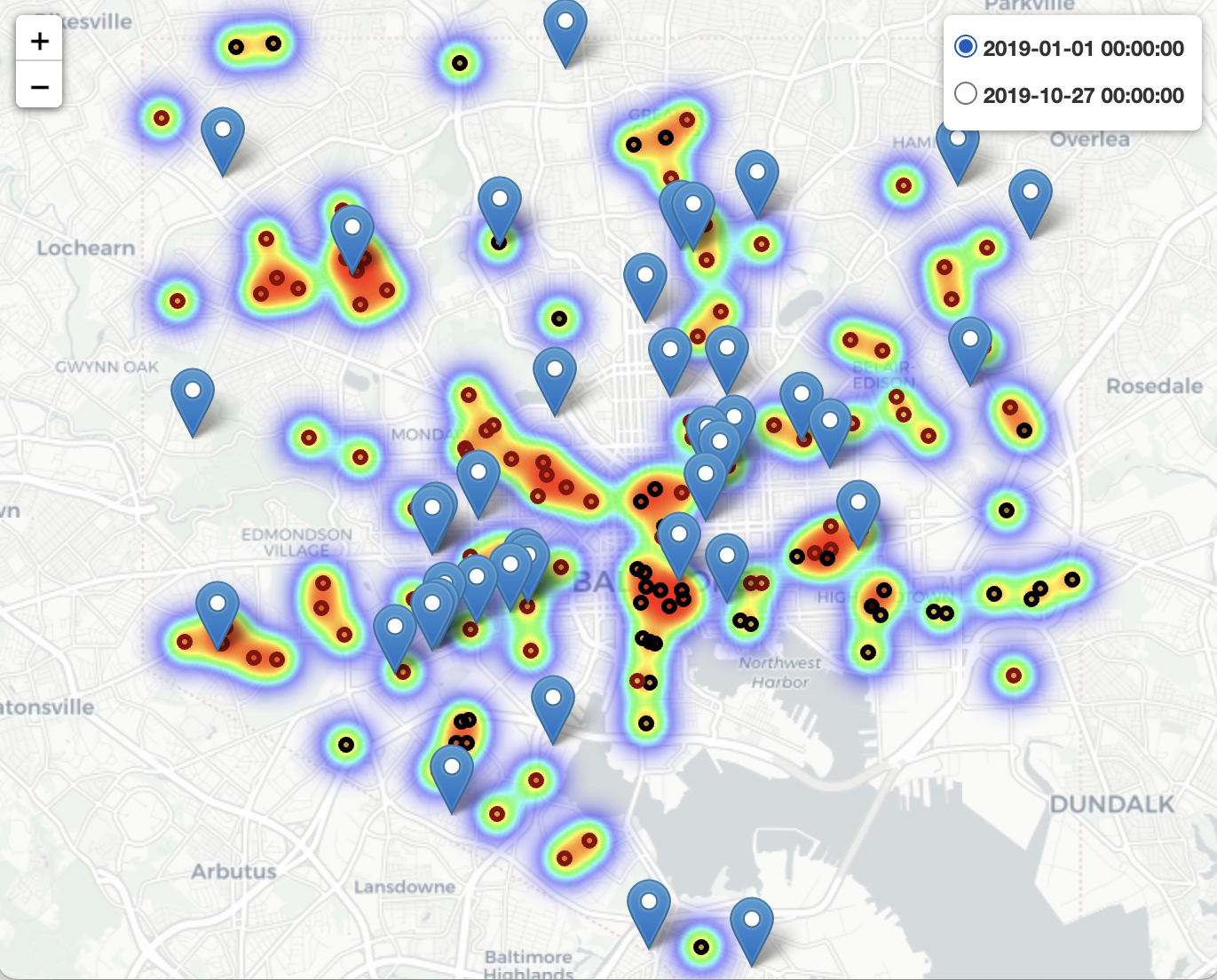}\\[-0.2em]
    \small Day 1
  \end{minipage}

  \vspace{0.75em}
  \begin{minipage}{0.32\textwidth}\centering
    \includegraphics[width=\linewidth]{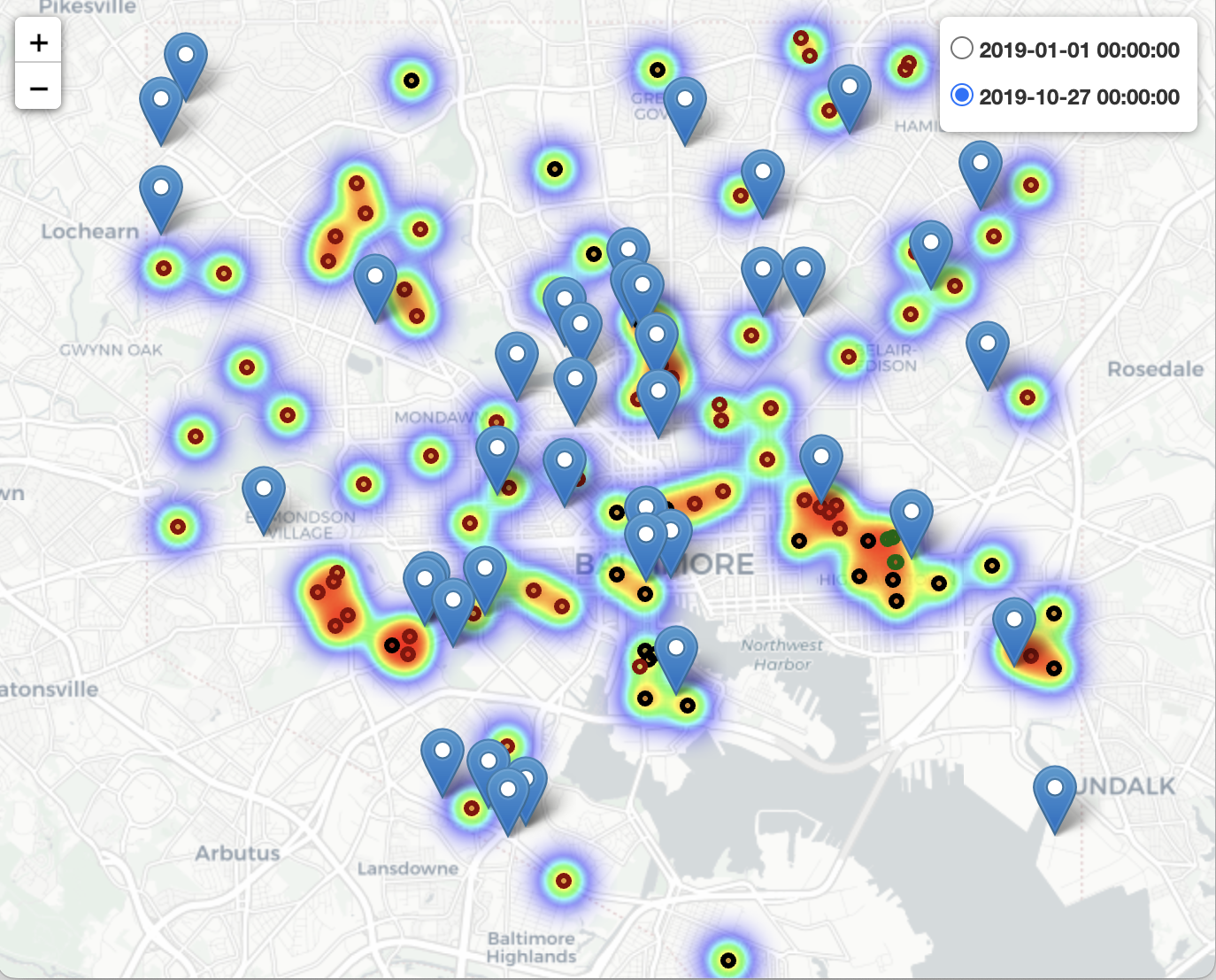}\\[-0.2em]
    \small Day 300
  \end{minipage}
  \hfill
  \begin{minipage}{0.32\textwidth}\centering
    \includegraphics[width=\linewidth]{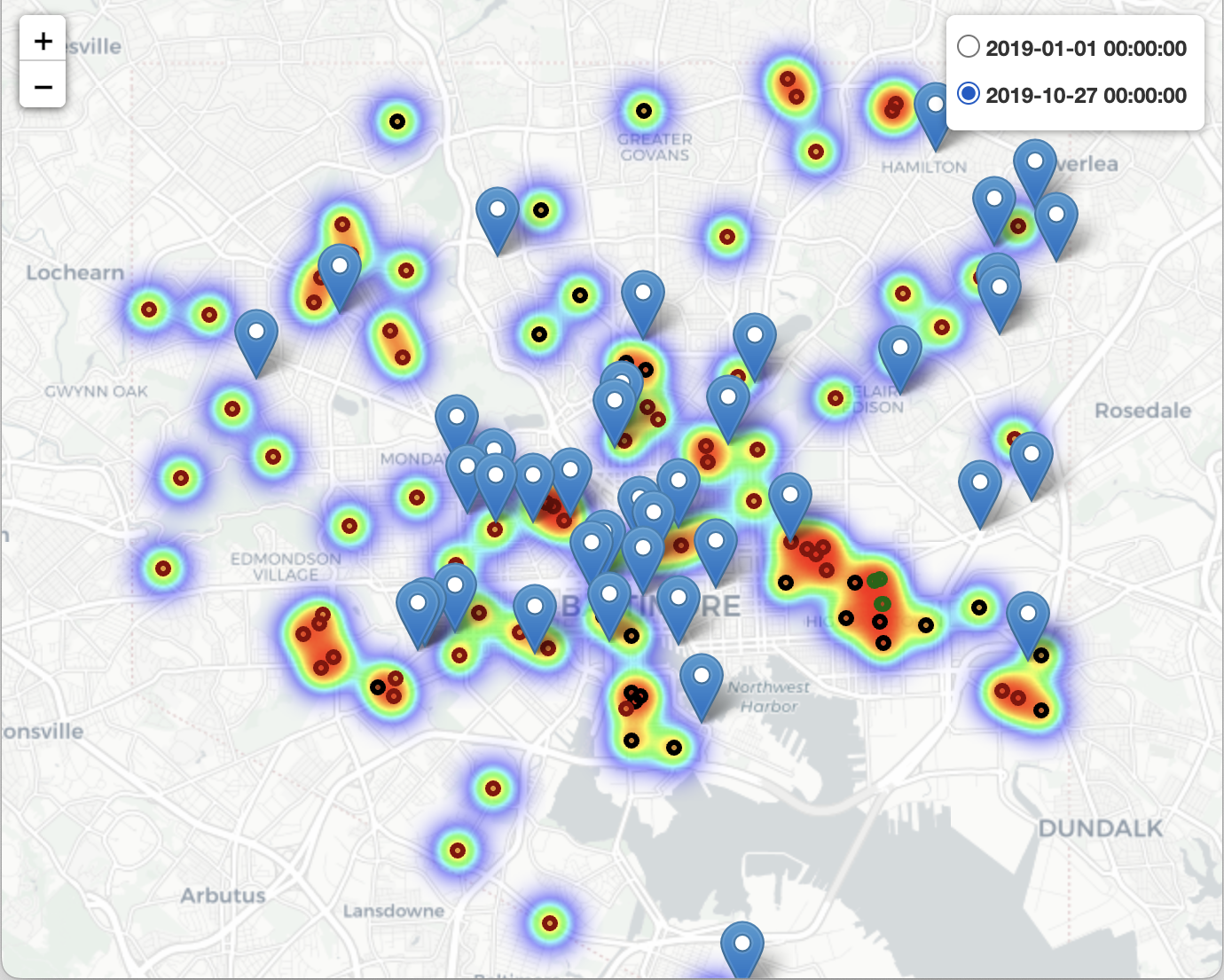}\\[-0.2em]
    \small Day 300
  \end{minipage}
  \hfill
  \begin{minipage}{0.32\textwidth}\centering
    \includegraphics[width=\linewidth]{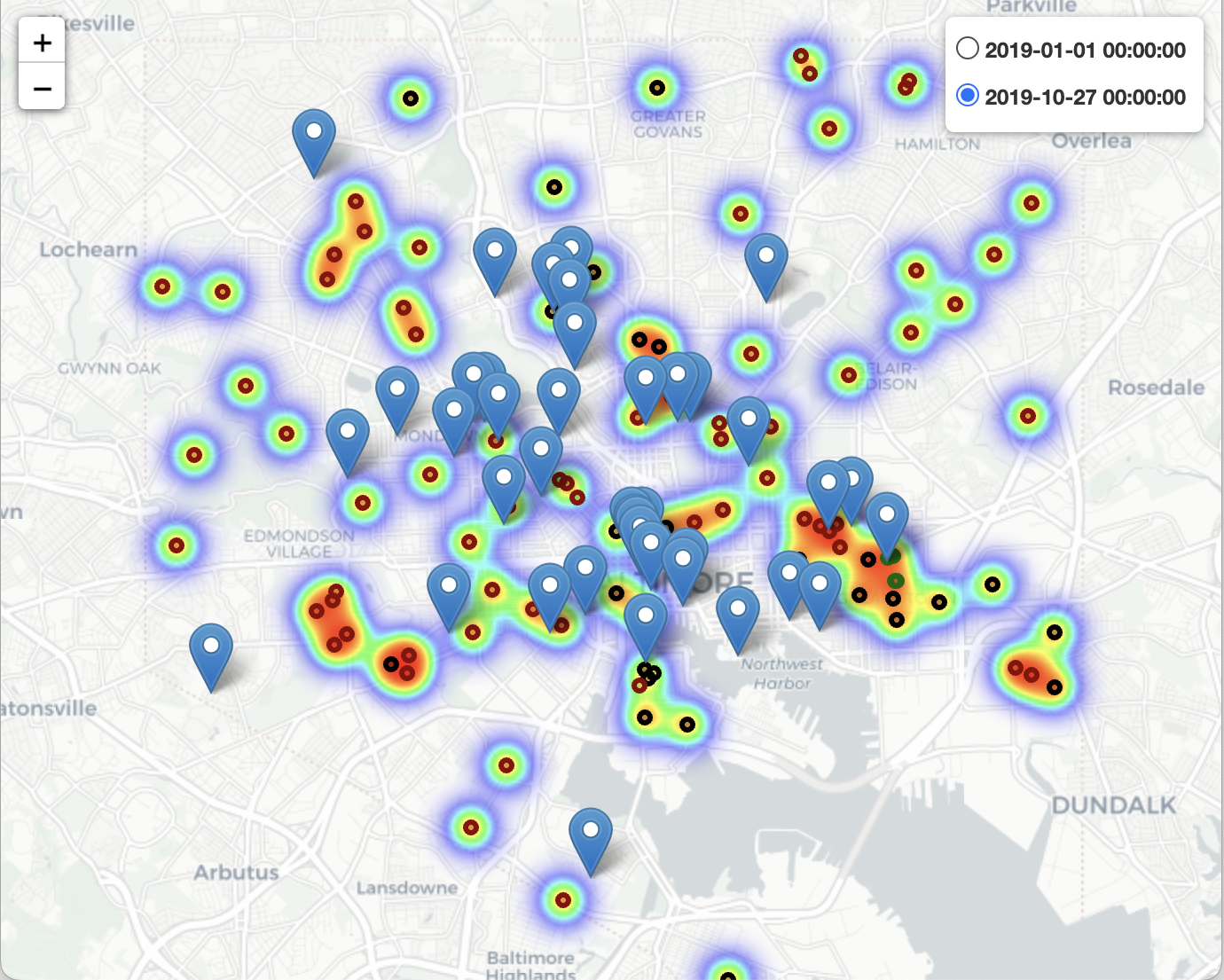}\\[-0.2em]
    \small Day 300
  \end{minipage}

  \caption{Simulation maps comparing three algorithms (columns). 
  T\jf{he t}op row shows police distribution on Day 1, while the bottom row shows distribution on day 300. All were simulated using report-probability $0.4$ and $40$ officers.}
  \label{maps_day1_300}
\end{figure}

In this section we describe the results of our experiments\jf{. }
The results dataset and code for the analysis are available online
.\footnote{\url{https://github.com/saminsemsar/Data_Analysis_Portfolio/tree/main/PredictivePolicing}}
The results are organized to address a series of guiding questions aimed at examining key aspects of our research objectives:
\begin{itemize}
    \item Which algorithm is fairer or more accurate in our 
    \jf{specific} location and time period?
    \item Do different systems react differently to feedback loops?
    \item Is the bias model-driven or data-driven?
    \item 
    \jf{H}ow 
    \jf{do} the police concentration and distribution change over the course of the simulation in different scenarios?
\end{itemize}
Before presenting the findings, we begin with a map-based visualization to provide an intuitive overview of how police were distributed on the first and last day of simulations in a\jf{n example} scenario 
\jf{(40 police officers, 300 days)}(\cref{maps_day1_300}). The red circles are crime spots in Black neighborhoods, while the Black circles are in White neighborhoods. The blue markers are police locations. 
It can be observed that police are allocated more evenly among neighborhoods on the first day of allocation compared to the last day (about a year later), as shown in~\cref{maps_day1_300}\jf{, illustrating the feedback loop phenomenon~\cite{lum2016predict}}. 
We now present the detailed findings, organized by themes corresponding to the guiding questions listed above.
\begin{enumerate}[label=\textbf{\arabic*.}]
    \item \textbf{Comparative Fairness and Accuracy}
    
    This subsection compares short-term KDE, Long-term KDE and PredPol in terms of overall fairness and accuracy within the Baltimore study region and time period. The findings summarize average and aggregate performance across scenarios, including whether the improvements in fairness affected accuracy.
    \begin{enumerate}[label=\textbf{Finding~\arabic{enumi}.\arabic{enumii}}]
        \item {\textbf{PredPol was the Most Accurate at the Beginning}}
        
        In 50\% of scenarios PredPol had the highest coverage accuracy compared to the other two on the simulation's first day. However, on the last day of the simulation, long-term KDE became the most accurate in 75\% of the scenarios\jf{, cf.} ~\cref{AlgoCompare_AvgPolShareblackMinusWhite_slope}.
        \begin{figure}[H]
            \begin{flushright}
                \begin{minipage}{0.85\linewidth}
                    \includegraphics[width=\linewidth, height=.95\textwidth]{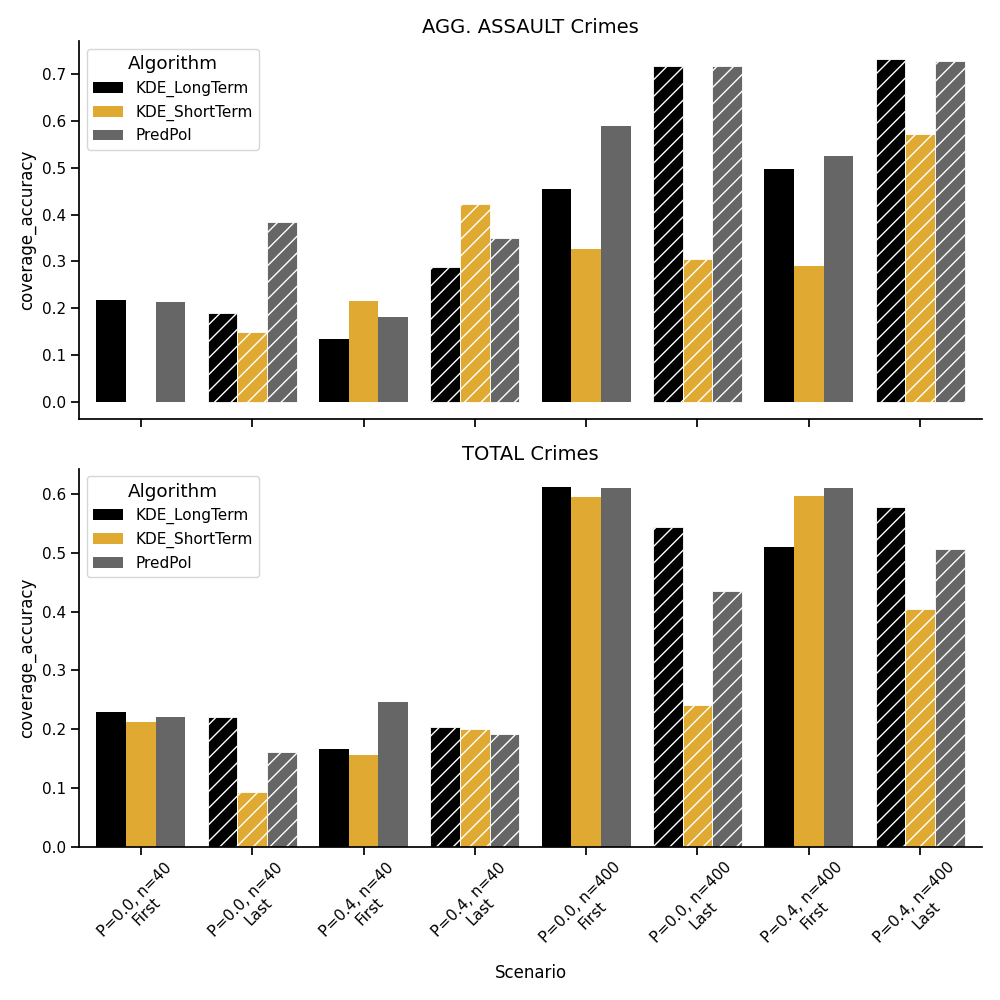}
                    \caption{First day vs. last day coverage accuracy of each algorithm over different scenarios.}
                    \label{AlgoCompare_AvgPolShareblackMinusWhite_slope}
                \end{minipage}
            \end{flushright}
        \end{figure}

        \item {\textbf{PredPol was Generally the Most Accurate Algorithm.}}
        
        PredPol was the most accurate algorithm in 87.5\% of scenarios and the second-most accurate in the rest. 
        \jf{Conversely,} short-term KDE was the least accurate in 75\% of scenarios (\cref{AlgorithmCompare_avg_accuracy}).
        \begin{figure}[H]
            \begin{flushright}
                \begin{minipage}{0.85\linewidth}
                    \includegraphics[width=\linewidth, height=.6\textwidth]{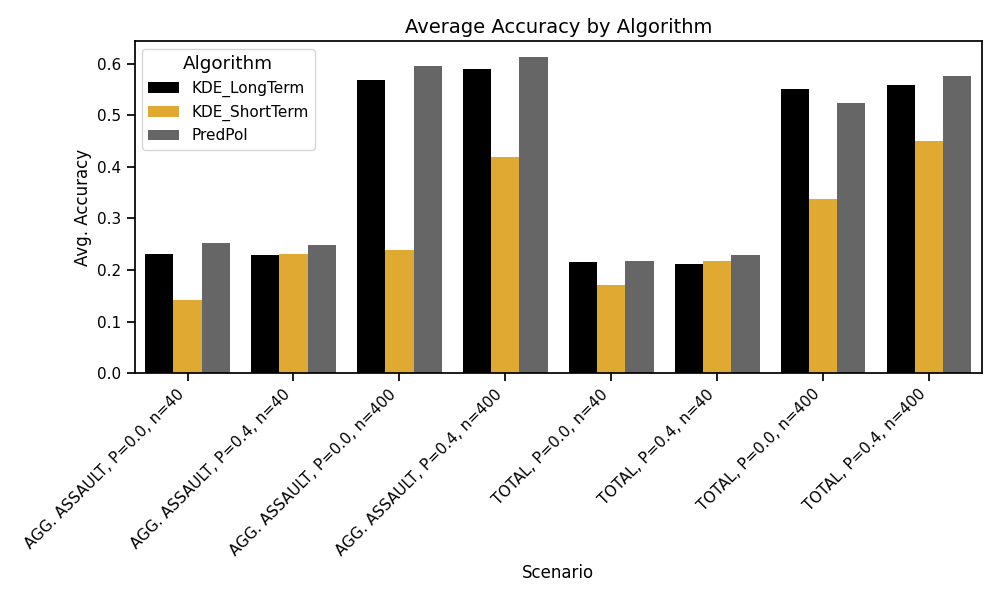}
                    \caption{Average accuracy by scenario and algorithm.}
                    \label{AlgorithmCompare_avg_accuracy}
                \end{minipage}
            \end{flushright}
        \end{figure}

        \item {\textbf{PredPol was the Most Racially Fair, While Long-Term KDE was the Least}.}
        
        PredPol was the most racially fair model in terms of average police share in 62.5\% of scenarios. Long-term KDE, on the other hand, was the least racially fair model in 75\% of scenarios (\cref{AlgoCompare_avg_racial_fairnessGap_avgPolShare}, and~\cref{Race Average Police-Share}).
        In terms of equality of police-share to crime-share ratio, PredPol was the most racially fair in 75\% of scenarios, and long-term KDE was the least racially fair in 87.5\% of scenarios (\cref{AlgoCompare_avg_Racial_Fgap_PCR}).
        \begin{figure}[H]
            \begin{flushright}
                \begin{minipage}{0.85\linewidth}
                    \includegraphics[width=\linewidth, height=.6\textwidth]{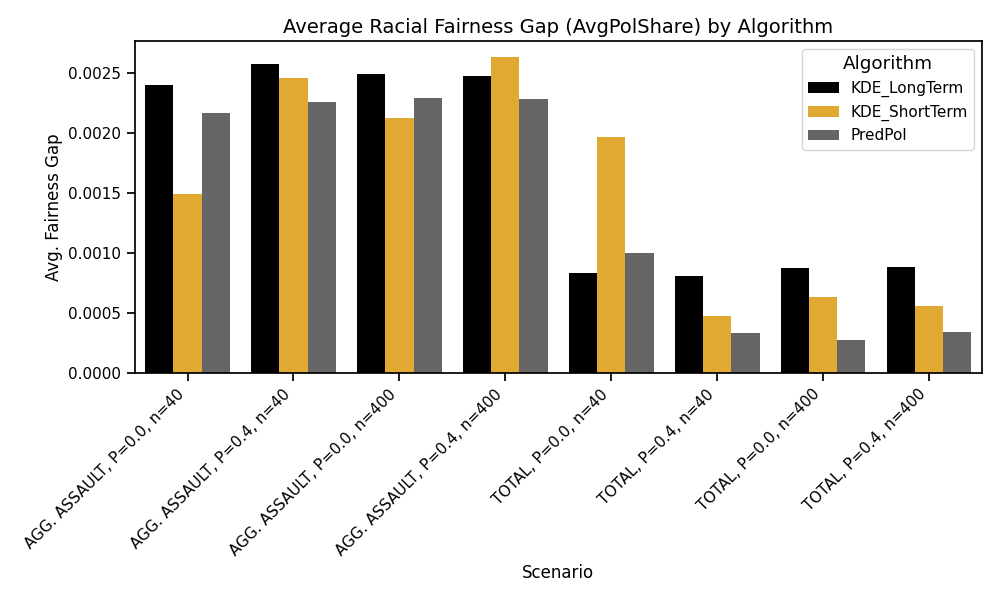}
                    \caption{Average racial fairness-gap based on average police-share, by scenario and algorithm\jf{.}}
                    \label{AlgoCompare_avg_racial_fairnessGap_avgPolShare}
                \end{minipage}
            \end{flushright}
        \end{figure}
        \begin{figure}[H]
            \begin{flushright}
                \begin{minipage}{0.85\linewidth}
                    \includegraphics[width=\linewidth, height=.6\textwidth]{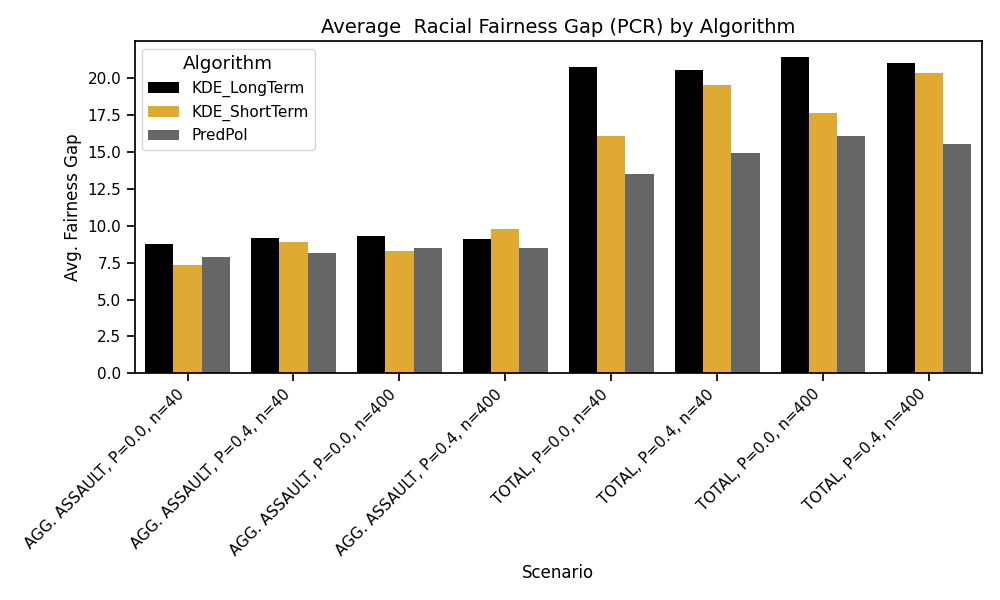}
                    \caption{Average racial fairness-gap based on police-to-crime ratios, by scenario and algorithm\jf{.}}
                    \label{AlgoCompare_avg_Racial_Fgap_PCR}
                \end{minipage}
            \end{flushright}
        \end{figure}
        
        \begin{figure}[H]
            \begin{flushright}
                \begin{minipage}{0.85\linewidth}
                    \subfloat[\centering Long-Term KDE]{
                      \includegraphics[width=0.95\linewidth, height=.55\textwidth]{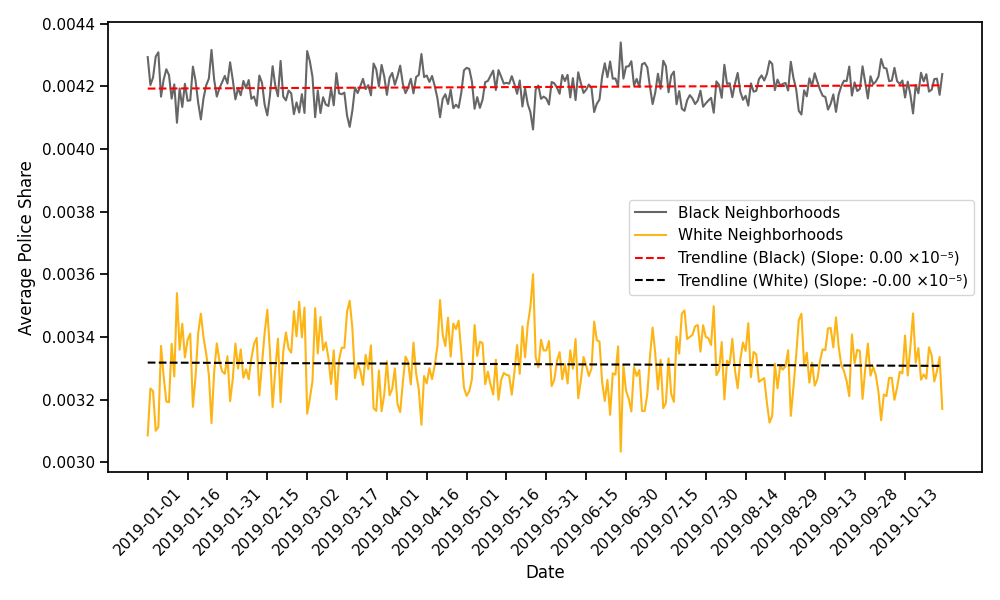}
                      \label{A}
                    }
                    \hfill
                    \subfloat[\centering Short-Term KDE]{
                      \includegraphics[width=0.95\linewidth, height=.55\textwidth]{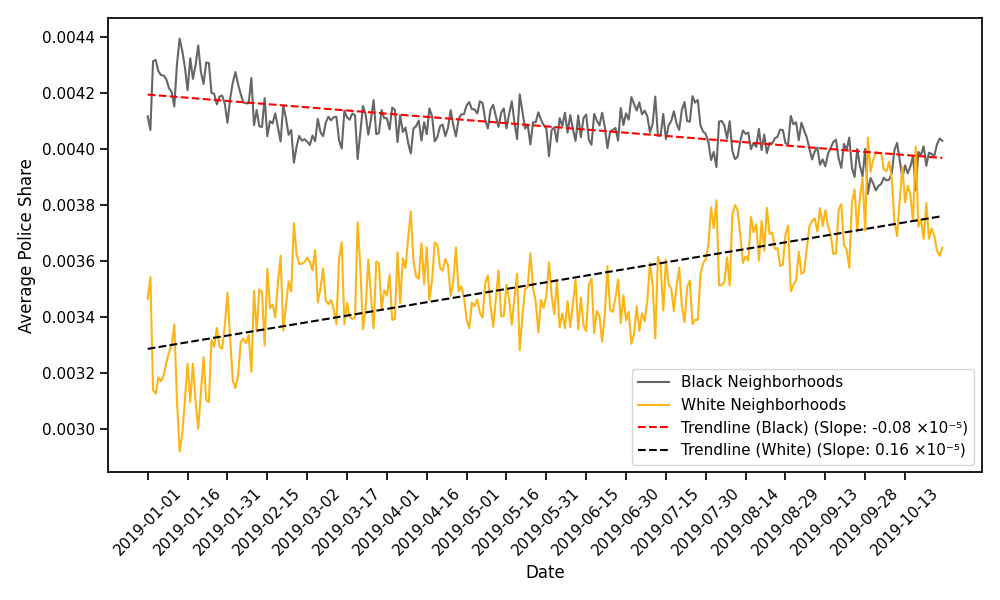}
                      \label{B}
                    }
                    \hfill
                    \subfloat[\centering PredPol]{
                      \includegraphics[width=0.95\linewidth, height=.55\textwidth]{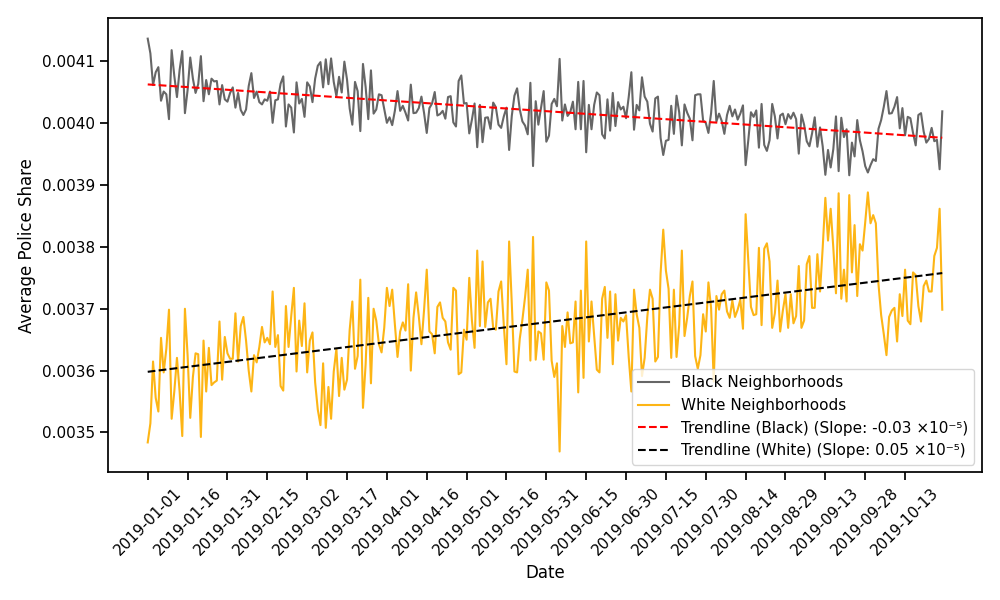}
                      \label{C}
                    }
                    \caption{Models' behavior based on average police-share assigned to Black and White neighborhoods over the days of the simulation, in a scenario example (distributing 400 officers, using all crime-records with crime report probability of 0.4)\jf{.}}
                    \label{Race Average Police-Share}
                \end{minipage}
            \end{flushright}
        \end{figure}
 
        \item {\textbf{PredPol was the Most Fair in Regards to Neighborhood-level Fairness, Followed by Long-term KDE, While Short-term KDE was the Least Neighborhood-level Fair Model.}}
        
        Based on both equality of average police share and equality of average police-to-crime ratio, in 75\% of scenarios PredPol and in the other 25\% long-term KDE were the most neighborhood-level fair models. 
        Short-term KDE was the least neighborhood-level fair in 100\% of scenarios according to both metrics (\cref{AlgoCompare_avg_NeighborhoodLevel_fairnessGap_pGini}, and~\cref{AlgoCompare_avg_NeighborhoodLevel_fairnessGap_pcr}).
        
        \begin{figure}[H]
            \begin{flushright}
                \begin{minipage}{0.85\linewidth}
                    \includegraphics[width=\linewidth, height=.7\textwidth]{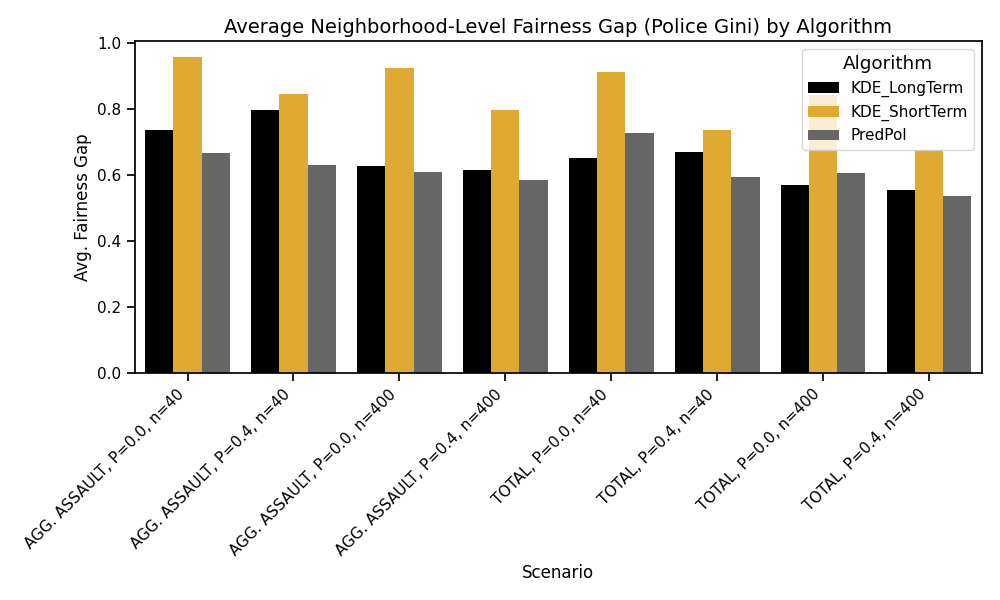}
                    \caption{Average neighborhood-level fairness-gap based on Gini coefficient of police number, by scenario and algorithm}
                    \label{AlgoCompare_avg_NeighborhoodLevel_fairnessGap_pGini}
                \end{minipage}
            \end{flushright}
        \end{figure}
        
        \begin{figure}[H]
            \begin{flushright}
                \begin{minipage}{0.85\linewidth}
                    \includegraphics[width=\linewidth, height=.7\textwidth]{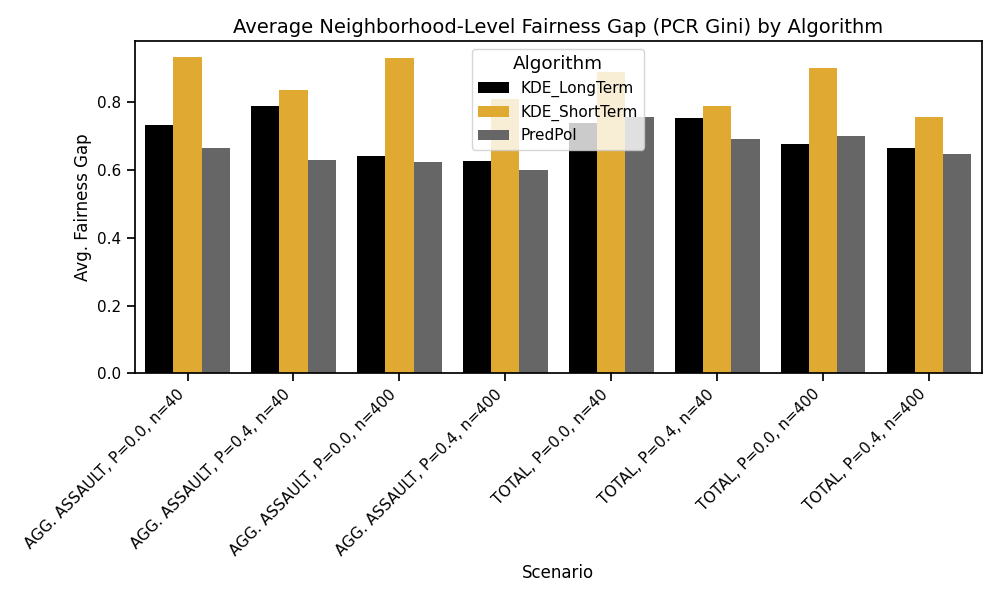}
                    \caption{Average neighborhood-level fairness-gap based on Gini coefficient of police-to-crime ratios, by scenario and algorithm}
                    \label{AlgoCompare_avg_NeighborhoodLevel_fairnessGap_pcr}
                \end{minipage}
            \end{flushright}
        \end{figure}
        
        \item {\textbf{Neighborhood-level Fairness Correlates with Accuracy.}}
        
        In~\cref{accuracy_vs_Neifairness}, it can be observed that higher neighborhood-level fairness gaps (lower fairness) were accompanied by lower accuracy. The accuracy–fairness correlation was clearer when the number of police officers allocated was higher.
        
        \begin{figure}[H]
            \begin{flushright}
              \begin{minipage}{0.85\linewidth}
                \subfloat[\centering Coverage accuracy vs. Neighborhood-Level Fairness Gap (PCR)]{
                  \includegraphics[width=\linewidth,height=.65\textwidth]{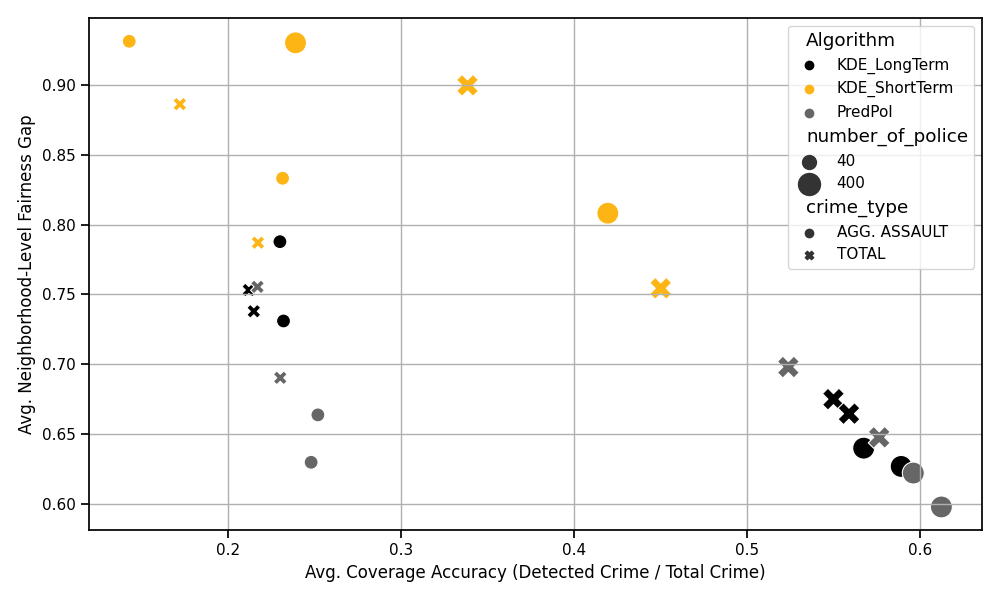}
                  \label{acc_vs_neipcr}
                }\par
                \subfloat[\centering Coverage accuracy vs. Neighborhood-Level Fairness Gap (Police Gini)]{
                  \includegraphics[width=\linewidth,height=.65\textwidth]{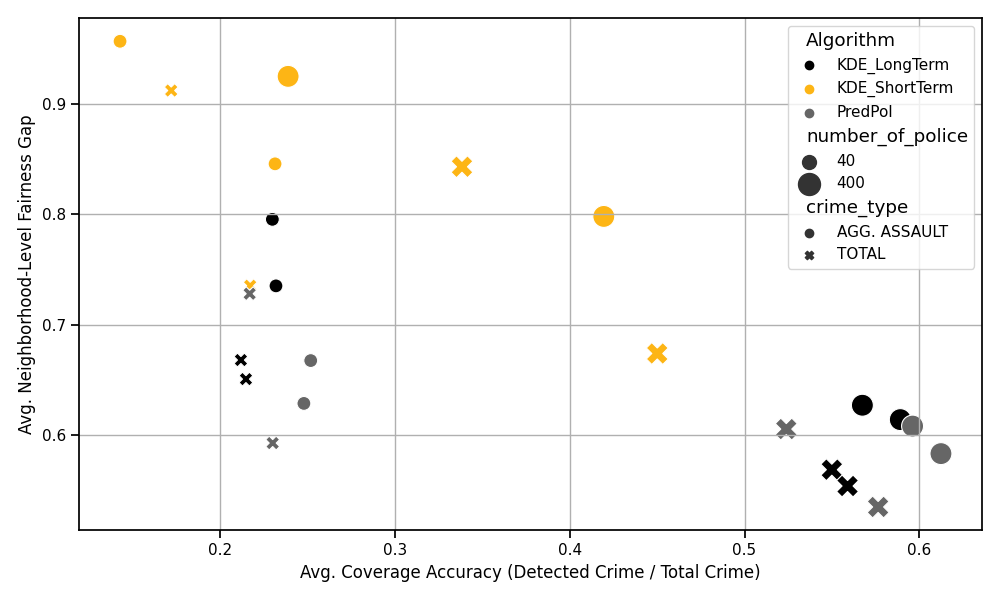}
                  \label{acc_vs_neigini}
                }\par\medskip
                
                \caption{Trade-offs between accuracy and neighborhood-level fairness across algorithms.
                Marker size indicates number of police. Marker shape indicates crime type.}
                \label{accuracy_vs_Neifairness}
              \end{minipage}
            \end{flushright}
        \end{figure}

        \item {\textbf{Racial Fairness Does not Correlate With Accuracy.}}
        
        Unlike neighborhood-level fairness, we 
        \jf{did not} observe any correlations between racial fairness and accuracy \jf{(}\cref{accuracy_vs_Racefairness}\jf{)}. All levels of racial fairness were observed over all levels of accuracy.
        
        \begin{figure}[H]
            \begin{flushright}
            \begin{minipage}{0.85\linewidth}
            \subfloat[\centering Coverage Accuracy vs. Racial Fairness Gap (PCR)]{
              \includegraphics[width=\linewidth,height=.6\textwidth]{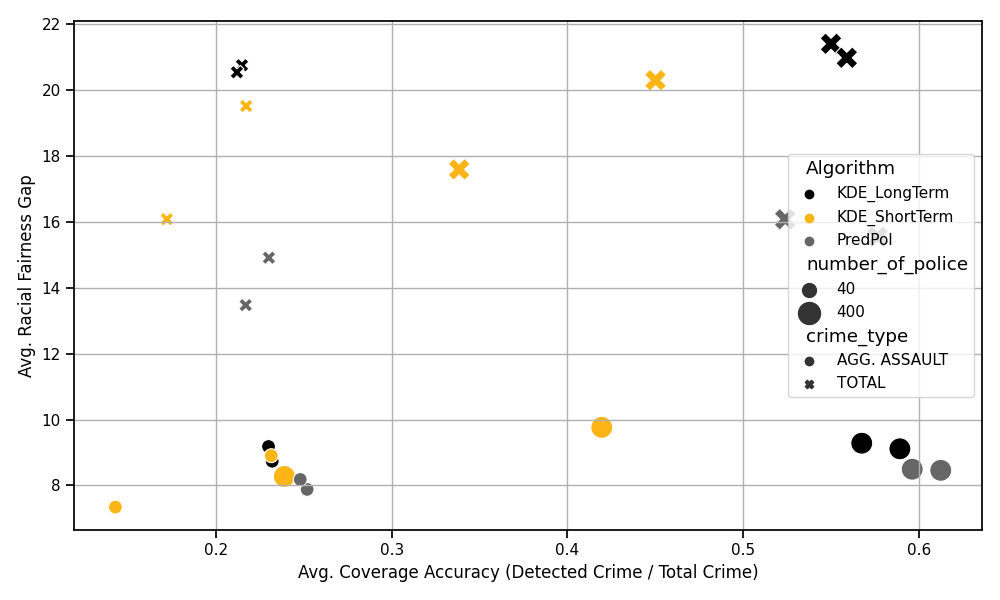}
              \label{acc_vs_racialpcr}
            }\par\medskip
            \subfloat[\centering Coverage accuracy vs. Racial Fairness Gap (Police Share)]{
              \includegraphics[width=\linewidth,height=.6\textwidth]{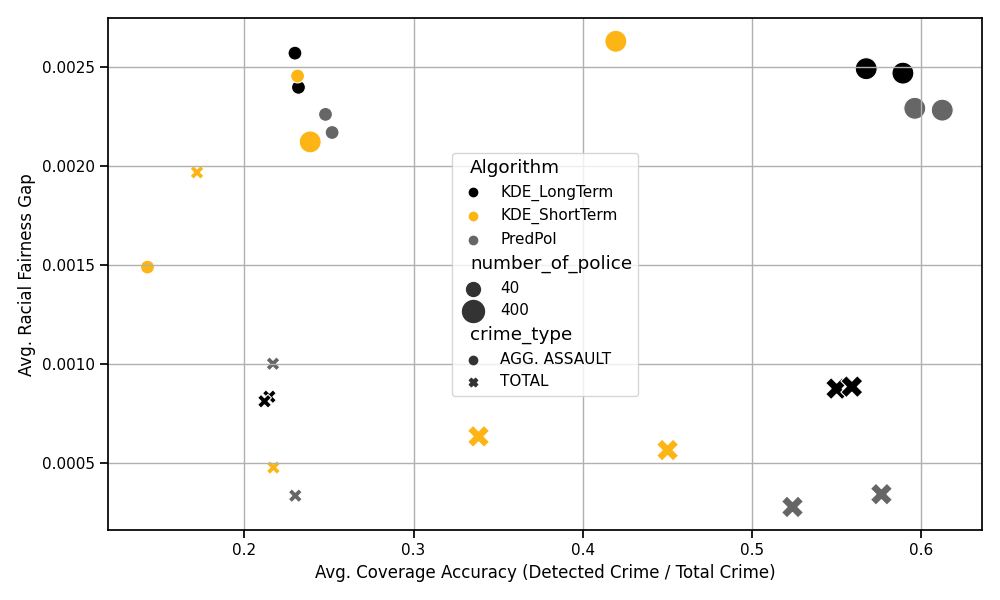}
              \label{acc_vs_racialshare}
            }
            \caption{Trade-offs between accuracy and racial fairness across scenarios. 
            Marker size indicates number of police. Marker shape indicates crime type.}
            \label{accuracy_vs_Racefairness}
            \end{minipage}
            \end{flushright}
        \end{figure}
    \end{enumerate}
    \item \textbf{Differential Responses to Feedback Loops}
    
    This subsection examines how different algorithms responded to feedback loops over time. These findings focus on temporal trends, including the direction and speed of change in fairness and accuracy, rather than average outcomes. 
    \begin{enumerate} [label=\textbf{Finding~\arabic{enumi}.\arabic{enumii}}]
        \item {\textbf{Long-Term KDE had the Slowest pace of Change in Most of the Scenarios, Especially for Neighborhood-level Fairness.}}
        
        Long-term KDE had the smallest slope of the trend line in:
        \begin{itemize}
            \item 50\% of scenarios, for racial fairness in terms of average police-share (\cref{AlgorithmCompare_racial_fairnessGap_slope_avgPolShare}),
            \begin{figure}[H]
                \begin{flushright}
                    \begin{minipage}{0.75\linewidth}
                        \includegraphics[width=\linewidth, height=.5\textwidth]{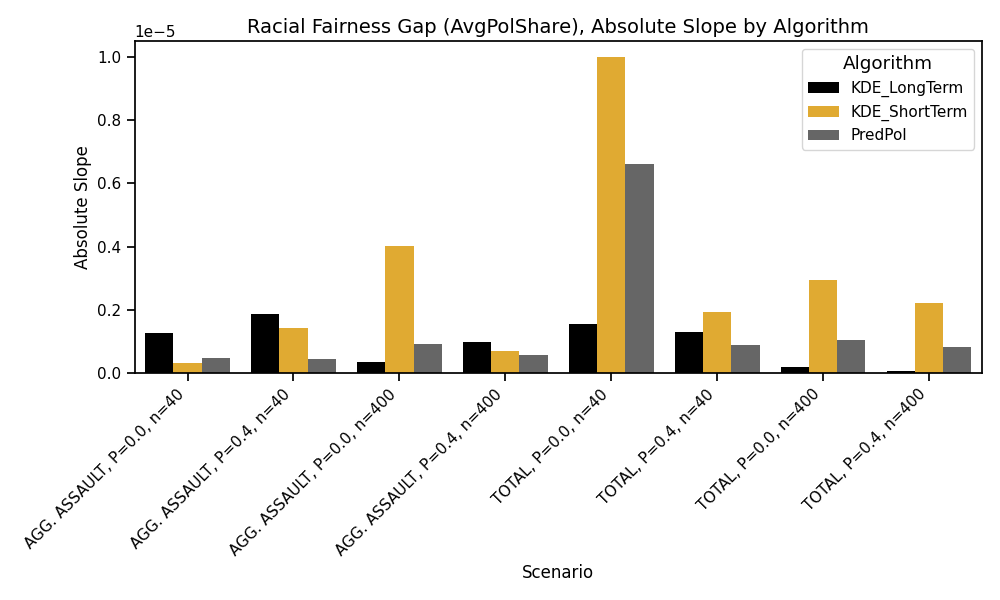}
                        \caption{Racial fairness gap (average police share), absolute slope by algorithm.}
                        \label{AlgorithmCompare_racial_fairnessGap_slope_avgPolShare}
                    \end{minipage}
                \end{flushright}
            \end{figure}
            \item 37.5\% of scenarios, for racial fairness in terms of police-to-crime proportionality (\cref{AlgorithmCompare_racial_fairnessGap_slope_pcr}),
            \begin{figure}[H]
                \begin{flushright}
                    \begin{minipage}{0.75\linewidth}
                        \includegraphics[width=\linewidth, height=.5\textwidth]{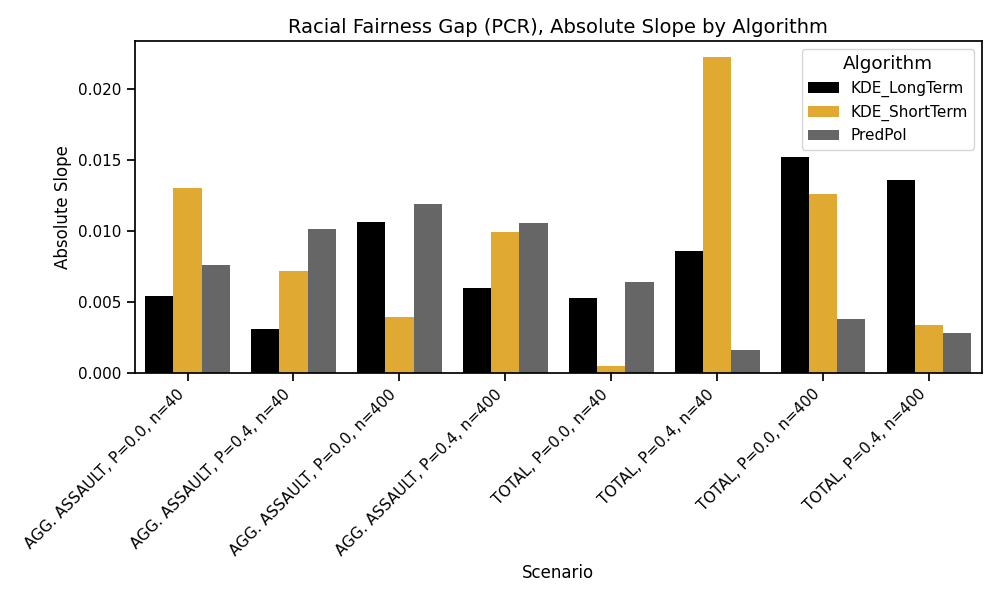}
                        \caption{Racial fairness gap (avg. police-to-crime ratio), absolute slope by algorithm.}
                        \label{AlgorithmCompare_racial_fairnessGap_slope_pcr}
                    \end{minipage}
                \end{flushright}
            \end{figure}
            \item 75\% of scenarios, for neighborhood-level fairness in terms of police-share equality (\cref{AlgorithmCompare_NeighborhoodLevel_fairnessGap_pGini_slope} and~\cref{PoliceGini}),
            \begin{figure}[H]
                \begin{flushright}
                    \begin{minipage}{0.75\linewidth}
                        \includegraphics[width=\linewidth, height=.5\textwidth]{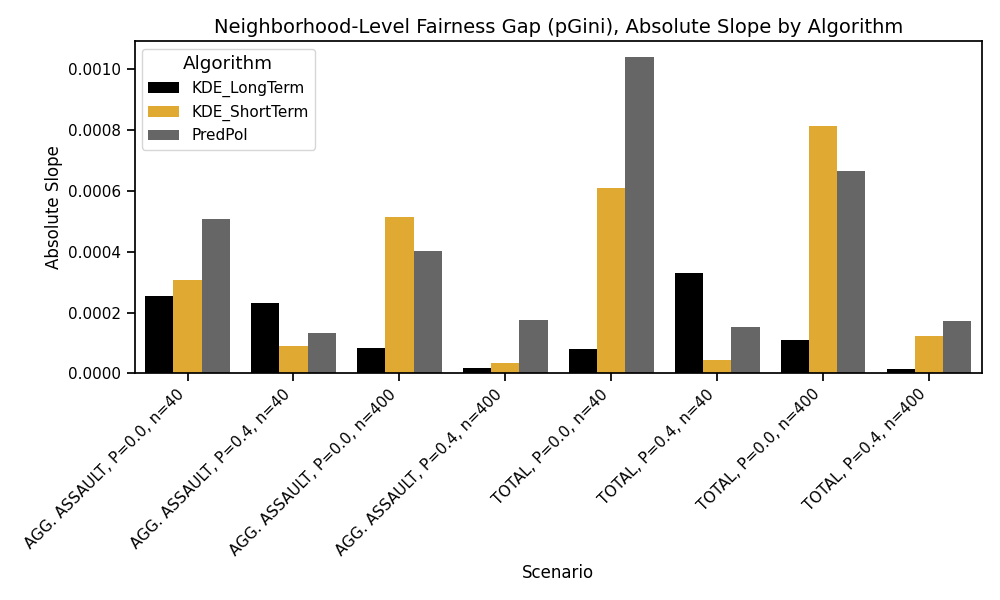}
                        \caption{Neighborhood-level fairness gap (average police share), absolute slope by algorithm.}
                        \label{AlgorithmCompare_NeighborhoodLevel_fairnessGap_pGini_slope}
                    \end{minipage}
                \end{flushright}
            \end{figure}
            
            \begin{figure}[H]
                \begin{flushright}
                    \begin{minipage}{0.8\linewidth}
                        \subfloat[\centering Long-Term KDE]{
                          \includegraphics[width=\linewidth, height=.6\textwidth]{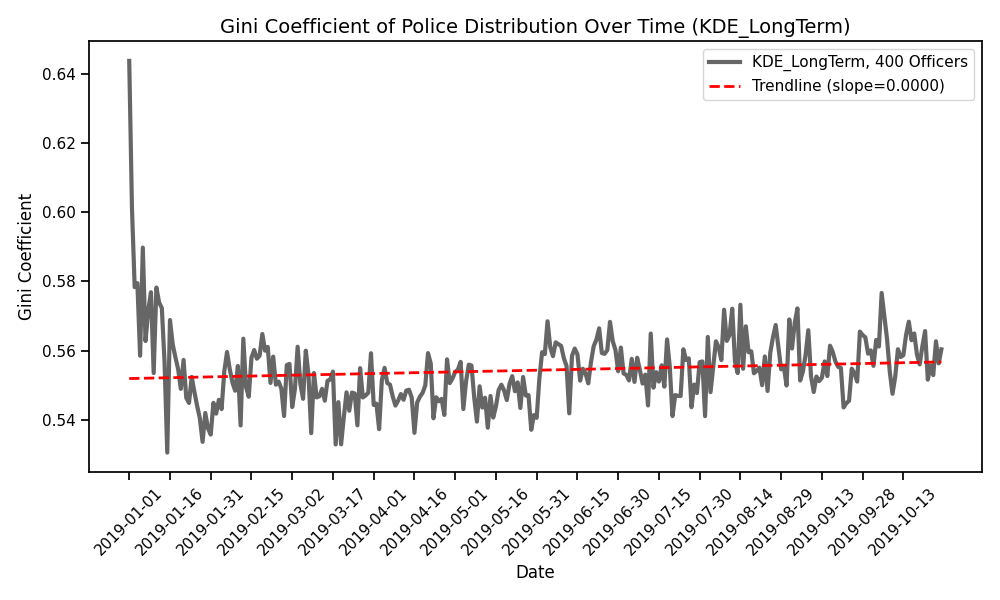}
                          \label{A}
                        }
                        \hfill
                        \subfloat[\centering Short-Term KDE]{
                        \includegraphics[width=\linewidth, height=.6\textwidth]{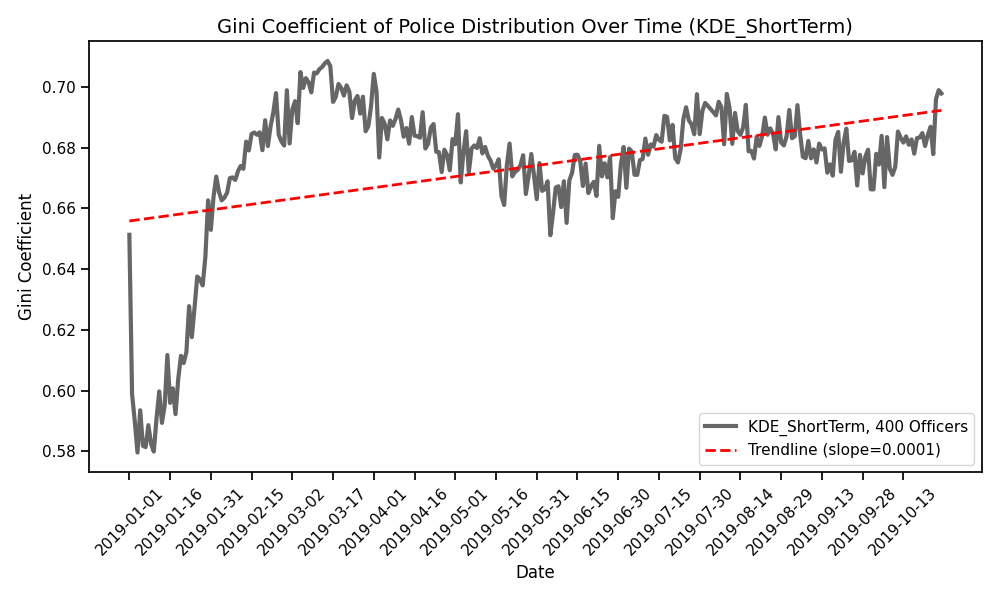}
                          \label{B}
                        }
                        \hfill
                        \subfloat[\centering PredPol]{
                          \includegraphics[width=\linewidth, height=.6\textwidth]{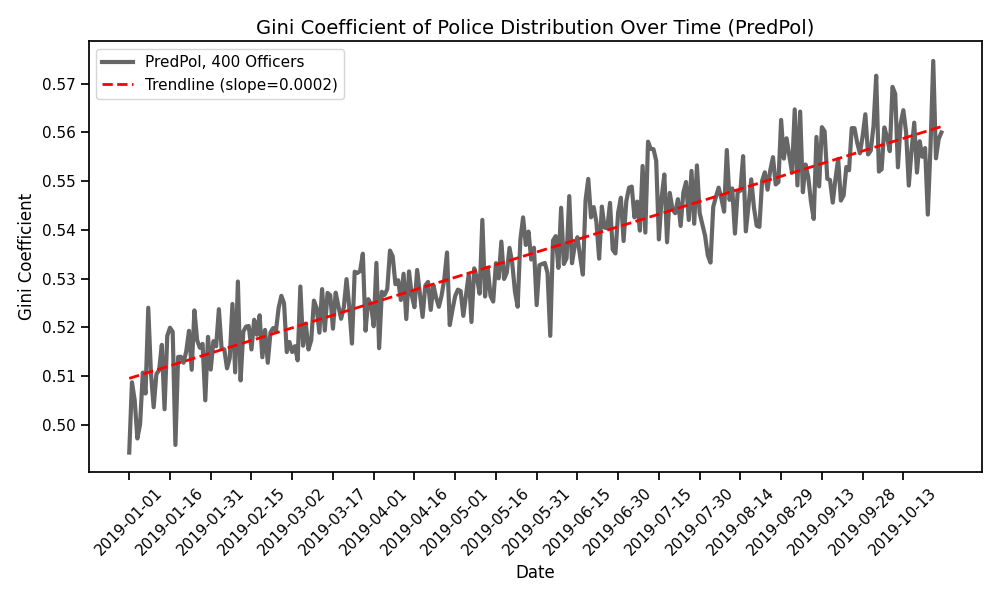}
                          \label{C}
                        }
                        \caption{Model's behavior based on neighborhood-level fairness-gap in terms of Gini coefficient of police numbers over the days of simulation, from a scenario example (distributing 400 officers, using all crime records with crime-report probability of 0.4)}
                        \label{PoliceGini}
                    \end{minipage}
                \end{flushright}
            \end{figure}
            \item 75\% of scenarios, for neighborhood-level fairness in terms of police-crime proportionality (~\cref{AlgorithmCompare_NeighborhoodLevel_fairnessGap_pcr_slope})
            \begin{figure}[H]
                \begin{flushright}
                    \begin{minipage}{0.85\linewidth}
                        \includegraphics[width=\linewidth, height=.6\textwidth]{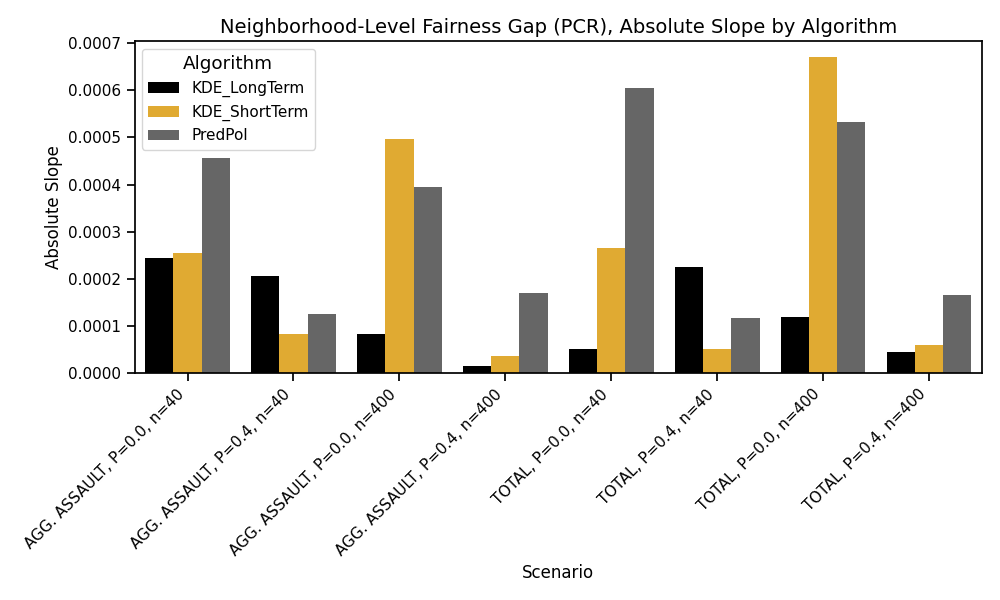}
                        \caption{Neighborhood-level fairness gap (police-to-crime ratio), absolute slope by algorithm.}
                        \label{AlgorithmCompare_NeighborhoodLevel_fairnessGap_pcr_slope}
                    \end{minipage}
                \end{flushright}
            \end{figure}
        \end{itemize}
        
        \item {\textbf{Short-term KDE had the Fastest Trend of Racial Bias Based on Average Police Share in Most of the Scenarios.}}
        
        Short-term KDE had the fastest trend of racial bias in 62.5\% of scenarios for equality of average police share (\cref{AlgorithmCompare_racial_fairnessGap_slope_avgPolShare}).
        Taking a closer look at individual scenarios, %
        \jf{a}lthough short-term KDE was ranked second in pace of neighborhood-level bias in most of the scenarios, its change in neighborhood-level fairness over the days showed a more volatile pattern compared to the other two models (\cref{PoliceGini}. To 
        \jf{see the} other scenarios, refer to the code~\cite{predictive_policing_code}).
        
        \item {\textbf{PredPol had the Fastest Trend of Neighborhood-level Bias and one of Racial Bias Metrics in Most of the Scenarios.}}
        
        We have already established that PredPol was generally the fairest model, both in terms of racial and neighborhood-level fairness. However, looking at the pace of change for each metric over different scenarios, 
        PredPol had a fast neighborhood-level fairness gap trend (\cref{AlgorithmCompare_NeighborhoodLevel_fairnessGap_pcr_slope},~\cref{AlgorithmCompare_NeighborhoodLevel_fairnessGap_pGini_slope}), and racial fairness trend in terms of police-to-crime ratio in most of the scenarios (\cref{AlgorithmCompare_racial_fairnessGap_slope_pcr}). \jf{This suggests that it is indeed vulnerable to bias from feedback loops, as previously reported by ~\cite{lum2016predict}.}

        \item {\textbf{Bias Amplification was Observed in a Higher Percentage of Scenarios for PredPol Compared to the Other Two Models.}}
        
        Although PredPol was fairer in most of the scenarios, the amplification of bias was seen in the trends in a higher percentage of scenarios for PredPol than the other two.
        PredPol's neighborhood-level fairness dropped in 100\% of scenarios, while the drop for short-term KDE and long-term KDE happened in 75\% and 37.5\%, respectively (\cref{AlgorithmCompare_WorseningOrImproving_NeighborhoodLevel_fairnessGap_pcr} and~\cref{AlgorithmCompare_WorseningOrImproving_NeighborhoodLevel_fairnessGap_pGini}). 
                
        \begin{figure}[H]
            \begin{flushright}
                \begin{minipage}{0.85\linewidth}
                    \includegraphics[width=\linewidth, height=.6\textwidth]{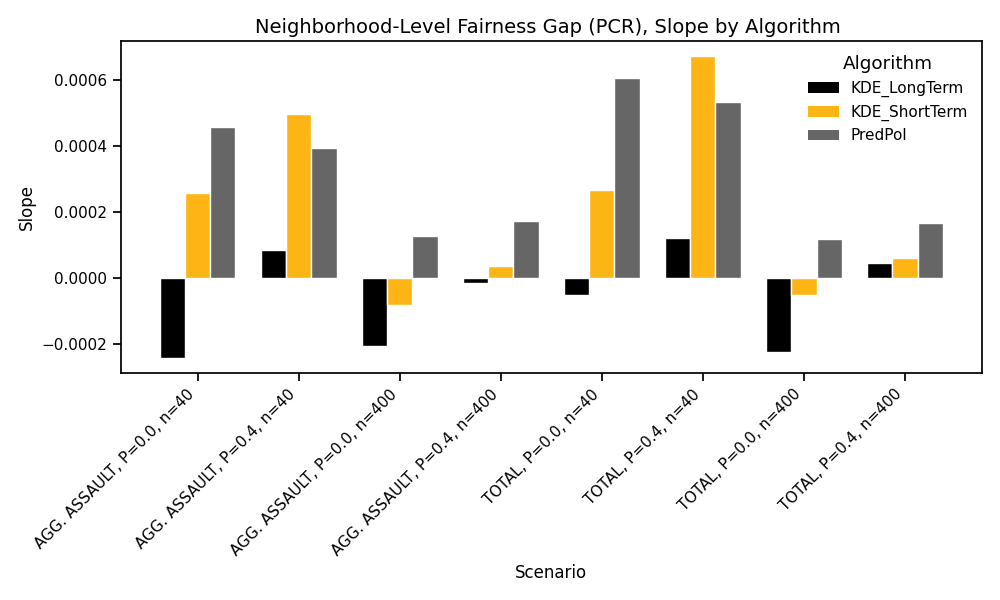}
                    \caption{Trend-line slope of neighborhood-level fairness gap based on Gini coefficient of police-crime proportionality over the days of simulation, across scenarios and algorithms.}
                    \label{AlgorithmCompare_WorseningOrImproving_NeighborhoodLevel_fairnessGap_pcr}
                \end{minipage}
            \end{flushright}
        \end{figure}

        \begin{figure}[H]
            \begin{flushright}
                \begin{minipage}{0.85\linewidth}
                    \includegraphics[width=\linewidth, height=.6\textwidth]{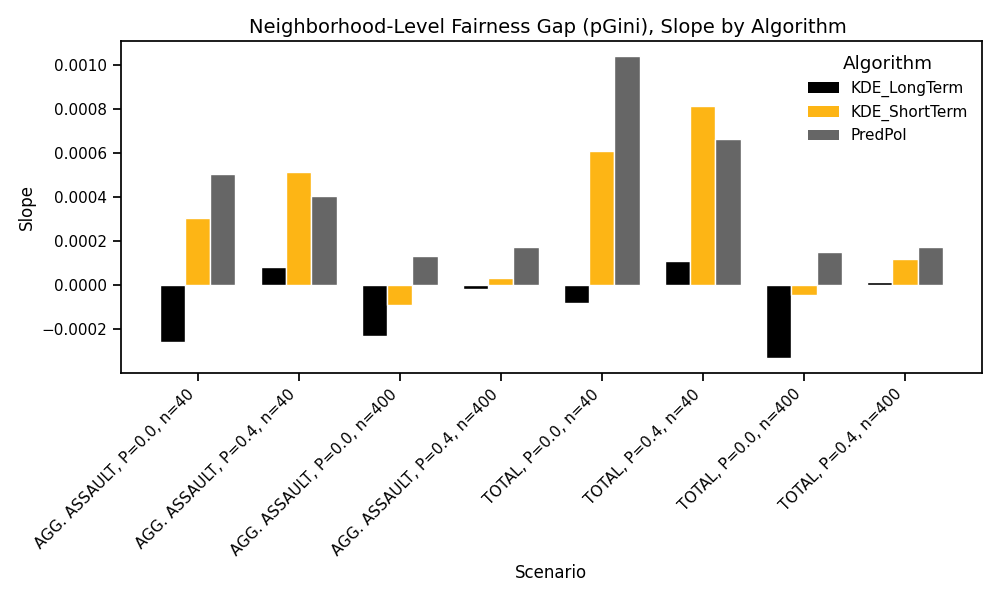}
                    \caption{Trend-line slope of neighborhood-level fairness gap based on Gini coefficient of police share over the days of simulation, across scenarios and algorithms.}
                    \label{AlgorithmCompare_WorseningOrImproving_NeighborhoodLevel_fairnessGap_pGini}
                \end{minipage}
            \end{flushright}
        \end{figure}
        
        The drop in racial fairness for PredPol happened in 75\% of scenarios for police-crime proportionality and 50\% of scenarios for average police share, while for short-term KDE these numbers were 50\% and 25\%, and for long-term KDE 75\% and 25\%, respectively (\cref{AlgorithmCompare_WorseningOrImproving_racial_fairnessGap_slope_PCR} and~\cref{AlgorithmCompare_WorseningOrImproving_racial_fairnessGap_slope_avgPolShare}). In \cref{AlgorithmCompare_WorseningOrImproving_racial_fairnessGap_slope_PCR} and \cref{AlgorithmCompare_WorseningOrImproving_racial_fairnessGap_slope_avgPolShare} the hatched or 
        \jf{striped} bars are those in which the race receiving focus at the start of the simulation contradicted with the race the trend was toward. This demonstrates how all the fairness gaps with negative slopes were the ones where the race focus at the beginning of the simulation contradicted the trend. 
        \jf{See, for instance the} models' behaviors in an example scenario in \cref{Race Average Police-Share}. 
        
        \begin{figure}[H]
            \begin{flushright}
                \begin{minipage}{0.85\linewidth}
                    \includegraphics[width=\linewidth, height=.6\textwidth]{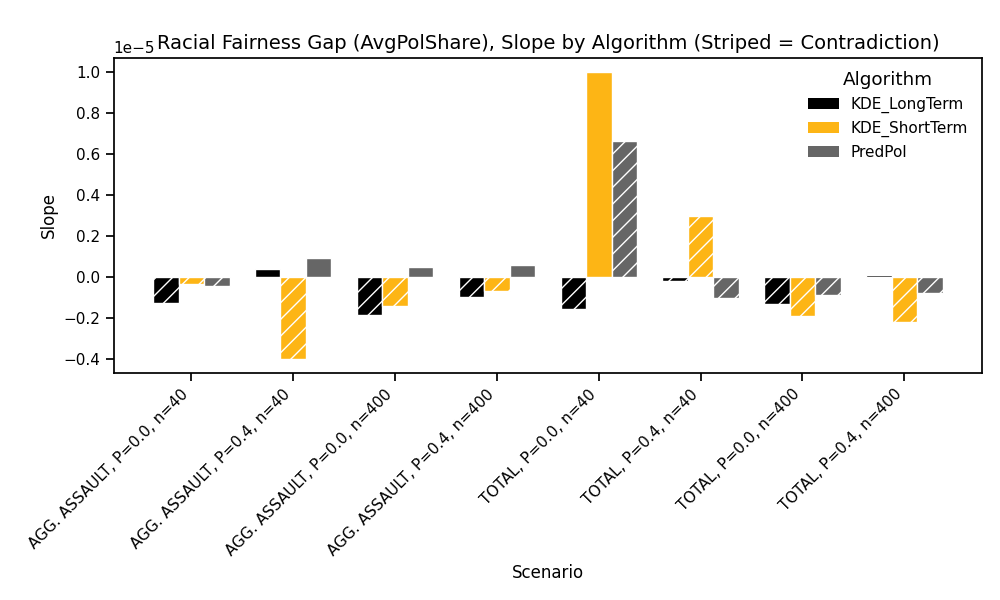}
                    \caption{Trend-line slope of racial fairness gap based on average police-share over the days of simulation, across scenarios and algorithms.}
                    \label{AlgorithmCompare_WorseningOrImproving_racial_fairnessGap_slope_avgPolShare}
                \end{minipage}
            \end{flushright}
        \end{figure}
        \begin{figure}[H]
            \begin{flushright}
                \begin{minipage}{0.85\linewidth}
                    \includegraphics[width=\linewidth, height=.6\textwidth]{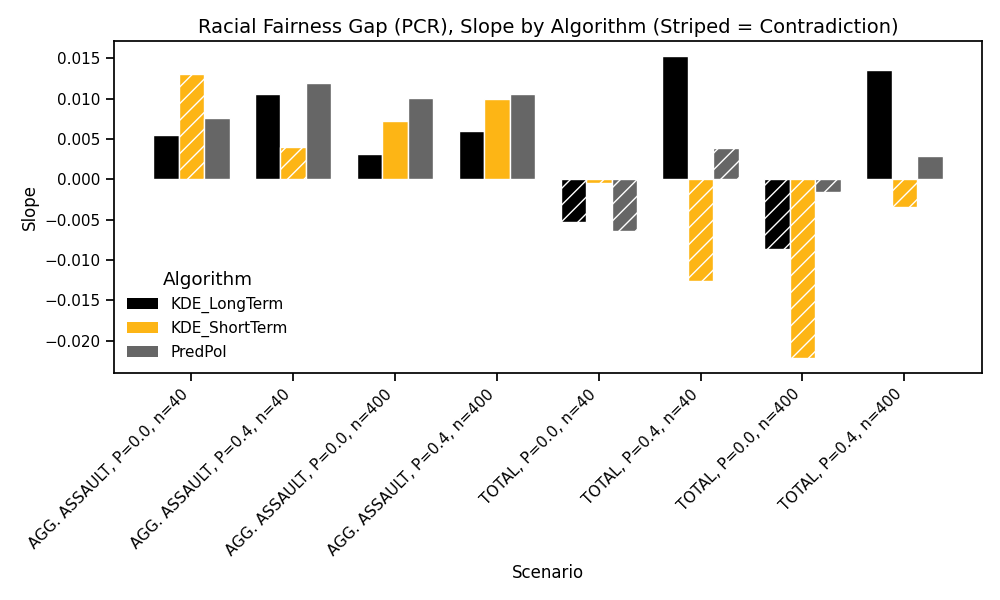}
                    \caption{Trend-line slope of racial fairness gap based on police-crime proportionality over the days of simulation, across scenarios and algorithms.}
                    \label{AlgorithmCompare_WorseningOrImproving_racial_fairnessGap_slope_PCR}
                \end{minipage}
            \end{flushright}
        \end{figure}

        \item {\textbf{Short-term KDE Experienced an Accuracy 
        \jf{Drop} in a Higher Percentage of Scenarios Compared to the Other two Models.}}
        
        Although we observed previously that neighborhood-level fairness and accuracy seemed to correlate to an extent, here we saw that short-term KDE, 
        \jf{n}ot the model with constant and highest speed of neighborhood-level fairness drop, experienced accuracy 
        \jf{drops} in 75\jf{\%} 
        of the scenarios. 
        \jf{The corresponding drop percentages were } 62.5\% for PredPol and 37.5\% for short-term KDE (\cref{AlgorithmCompare_WorseningOrImproving_accuracy_slope}).   
        \begin{figure}[H]
            \begin{flushright}
                \begin{minipage}{0.85\linewidth}
                    \includegraphics[width=\linewidth, height=.6\textwidth]{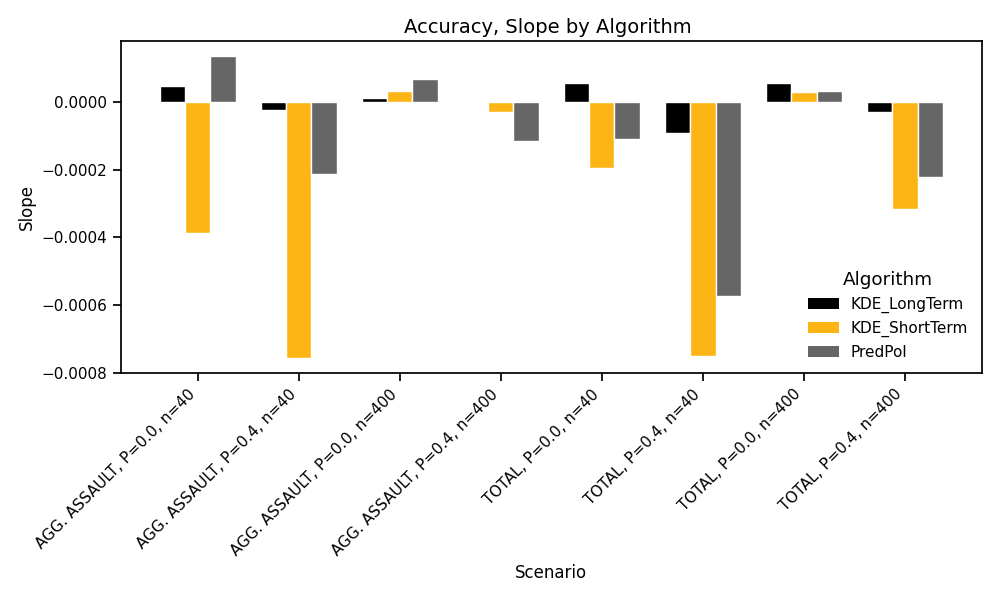}
                    \caption{Trend-line slope of accuracy over the days of simulation, across scenarios and algorithms.}
                    \label{AlgorithmCompare_WorseningOrImproving_accuracy_slope}
                \end{minipage}
            \end{flushright}
        \end{figure}

        \item {\textbf{Distribution Uniformity Didn't Guarantee the Defined Racial Fairness.}}
        
        While percentages of scenarios with a drop in neighborhood-level fairness ranged from 75\% to 100\% of scenarios for PredPol and short-term KDE, \jf{the} percentages of scenarios with a racial-fairness drop ranged from 25 to 75 (\cref{AlgorithmCompare_WorseningOrImproving_NeighborhoodLevel_fairnessGap_pcr},~\cref{AlgorithmCompare_WorseningOrImproving_NeighborhoodLevel_fairnessGap_pGini},~\cref{AlgorithmCompare_WorseningOrImproving_racial_fairnessGap_slope_avgPolShare},~\cref{AlgorithmCompare_WorseningOrImproving_racial_fairnessGap_slope_PCR}). For long-term KDE on the other hand, we saw racial fairness based on police-crime proportionality drop in 75\% of scenarios, while both neighborhood-level fairness metrics were worsening in 37.5\% of scenarios. 
        \jf{L}ong-term KDE's racial fairness, based on average police share, however, was only dropping in 25\% of scenarios. What this indicated was that the models with the least police distribution uniformity might stay racially fair in an area with a certain demographic map and crime records.

        \item {\textbf{Bias Typically Worsened over Time Except When Trend and Current State Contradicted.}}
        
        Looking across the scenarios, we observed a consistent pattern: in the cases where racial fairness appeared to improve unexpectedly, the demographic group receiving the greatest police attention at the start of the simulation was different from the group receiving the most attention at the end. In the case of Baltimore using all crime records, this focus at the end was surprisingly on White neighborhoods. This shift largely accounted for the observed improvement (see~\cref{AlgorithmCompare_WorseningOrImproving_racial_fairnessGap_slope_avgPolShare} and~\cref{AlgorithmCompare_WorseningOrImproving_racial_fairnessGap_slope_PCR}). For these contradictory cases—where the initial state and the direction of change do not align with bias amplification—the slope of the bias trend must have been sufficiently steep relative to the initial disparity for the trajectory to have crossed the parity point and begun moving toward amplification. In such situations, longer simulation horizons are required for this transition to appear.
        
        These apparent improvements hint at the possibility of seasonal cycles of bias reduction and subsequent amplification when simulations are run for extended durations. Changes in real-crime data distribution may arise from multiple underlying factors. Two such factors are illustrated below: 
        \begin{enumerate}
            \item Seasonality: any cyclic affecting variable could cause seasonal change in the distribution. An obvious one is the weather.
            \item Police policy change: changes in police distribution or enforcement practices could reduce  crime in a location or suppress some specific crime types affected by that policy, which induces a shift in crime distribution.
        \end{enumerate}
        Therefore, when real-crime data are used, and both reported and detected crimes contribute to the updates, the distribution of reported crimes may begin to favor a neighborhood that was not previously predicted to be high-risk. As officers gradually shift their attention from the earlier high-crime-rate areas toward this newly emergent hotspot, a temporary reduction in measured bias can occur. This is typically followed by renewed amplification once police resources become concentrated in the new focal area, until the reported-crime distribution eventually shifts again.
    \end{enumerate}
    \item\textbf{The Effects of Data Variation on Bias}
    
    This subsection examines how different crime data affected the fairness and accuracy outcomes across the algorithms. We compared scenarios using aggravated assault records with those using all crime records to analyze the change in fairness and accuracy.
    \begin{enumerate}[label=\textbf{Finding~\arabic{enumi}.\arabic{enumii}}]

        \item {\textbf{Expanding from Aggravated Assault to All-crimes, Increased Inequality of  Police-To-Crime Ratio and Reduced Accuracy for Long-term KDE and PredPol.}}
        
        When changing the crime type from aggravated assault to all crimes, the neighborhood-level gap based on the police-to-crime ratio and accuracy dropped in most of the scenarios (see \cref{CrimeTypeCompare_avg_NeighborhoodLevel_fairnessGap_pcr} and \cref{CrimeTypeCompare_avg_accuracy}). 
        \begin{figure}[H]
            \begin{flushright}
                \begin{minipage}{0.8\linewidth}
                    \includegraphics[width=\linewidth, height=.6\textwidth]{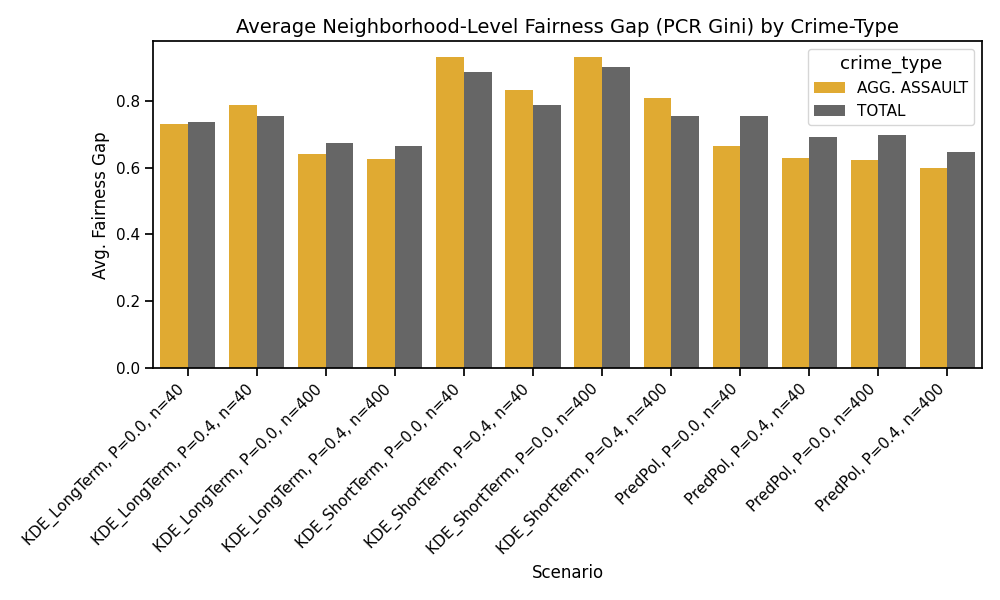}
                    \caption{Comparing average neighborhood-level fairness gaps (based on police-to-crime ratios) of similar scenarios with different crime data.}
                    \label{CrimeTypeCompare_avg_NeighborhoodLevel_fairnessGap_pcr}
                \end{minipage}
            \end{flushright}
        \end{figure}
        \begin{figure}[H]
            \begin{flushright}
                \begin{minipage}{0.8\linewidth}
                    \includegraphics[width=\linewidth, height=.6\textwidth]{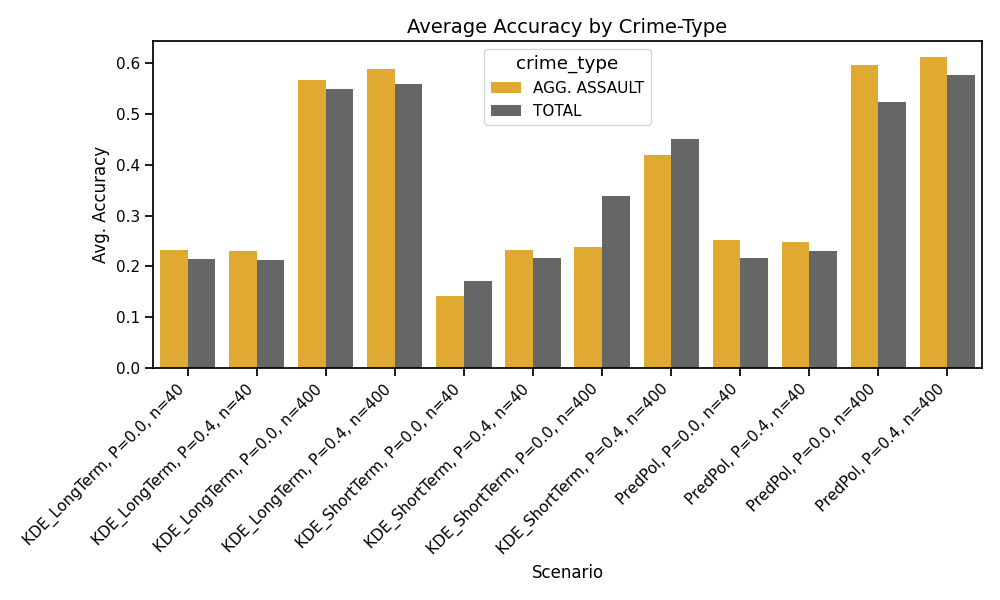}
                    \caption{Comparing average accuracy of similar scenarios with different crime data.}
                    \label{CrimeTypeCompare_avg_accuracy}
                \end{minipage}
            \end{flushright}
        \end{figure}
        
        \item {\textbf{Various Models Could have different Bias Outcomes When Applied to Different Data and Vice Versa.}}
        
        Looking at the slope of the trendline to determine which race is getting a higher average police share (\cref{CrimeTypeCompare_direction_avgPolShare}) or a higher average police-to-crime ratio (\cref{CrimeTypeCompare_direction_PCR}) as the simulation days pass, we observe that when feeding aggravated assault crime records to PredPol, the focus is more on Black neighborhoods, while using all crime records, the focus flips to White neighborhoods in most scenarios for both metrics. \jf{This is noteworthy since the debate around bias in predictive policing typically implicitly assumes that the algorithms are solely biased against Black neighborhoods, following the work of Lum and Isaac~\cite{lum2016predict}.}
        \begin{figure}[H]
            \begin{flushright}
                \begin{minipage}{0.87\linewidth}
                    \includegraphics[width=\linewidth, height=.63\textwidth]{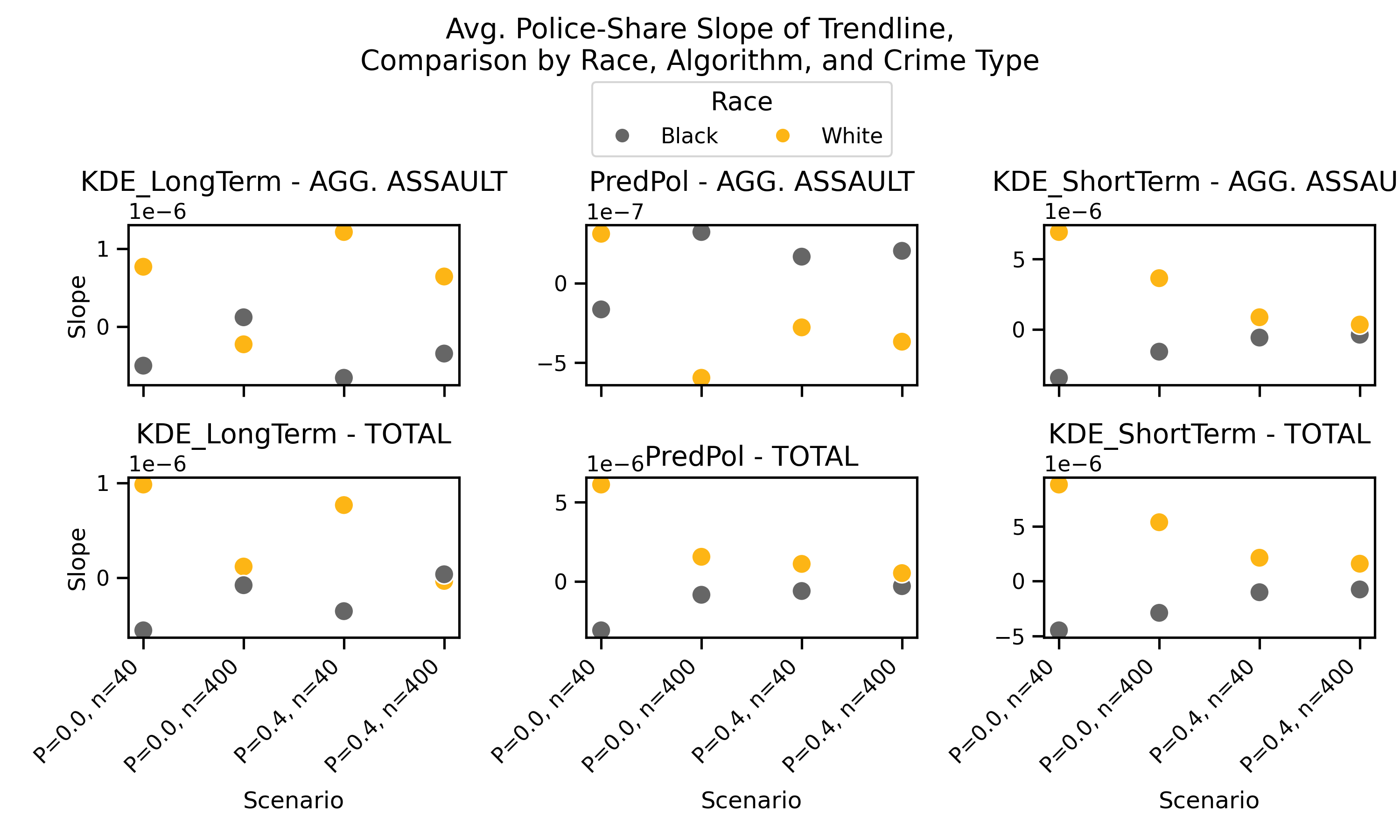}
                    \caption{Comparing Slope of Trendline of Black vs. White neighborhoods for the average police share across scenarios with different algorithms and crime data.}
                    \label{CrimeTypeCompare_direction_avgPolShare}
                \end{minipage}
            \end{flushright}
        \end{figure}  
        
        \begin{figure}[H]
            \begin{flushright}
                \begin{minipage}{0.87\linewidth}
                    \includegraphics[width=\linewidth, height=.63\textwidth]{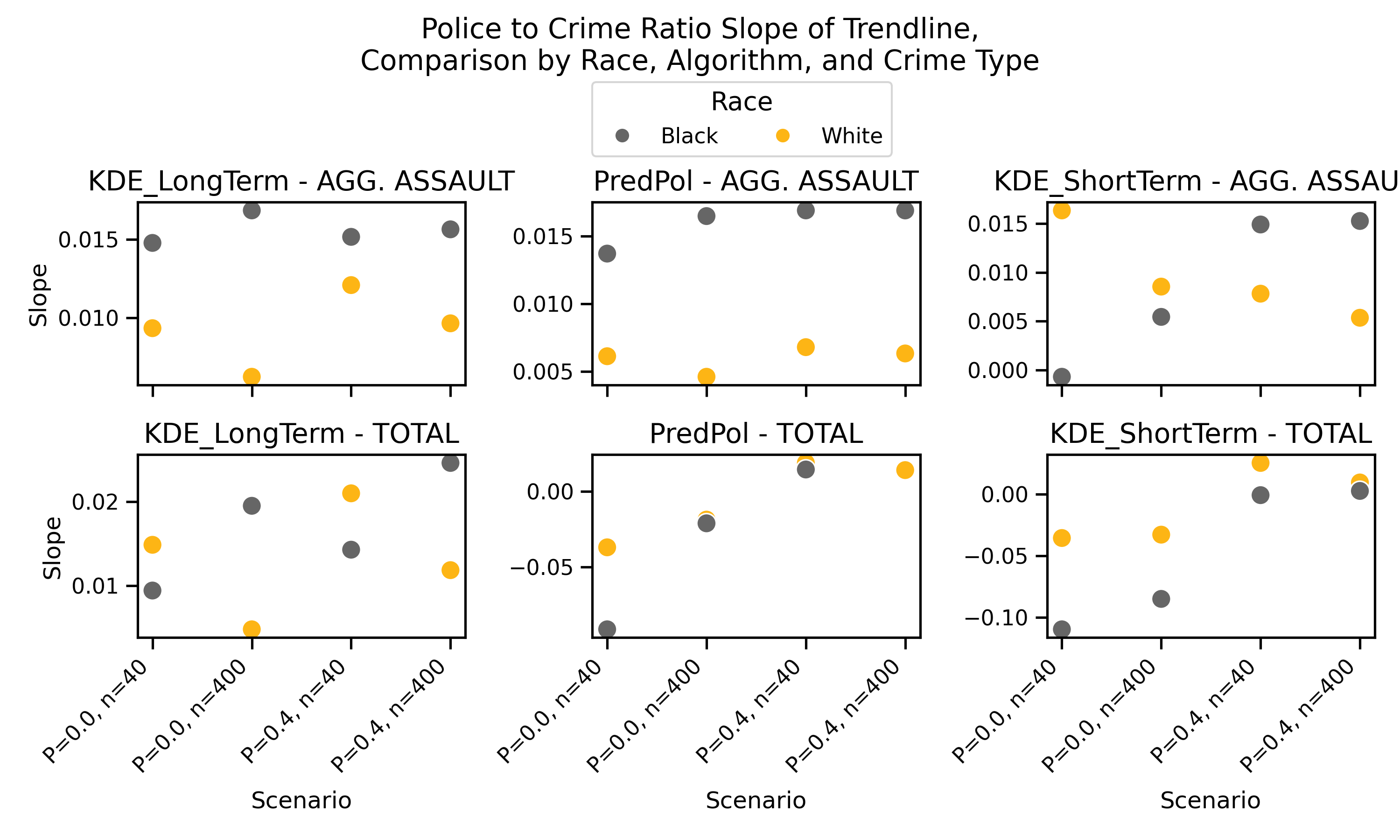}
                    \caption{Comparing Slope of Trendline of Black vs. White neighborhoods for the police-to-crime ratio across scenarios with different algorithms and crime data.}
                    \label{CrimeTypeCompare_direction_PCR}
                \end{minipage}
            \end{flushright}
        \end{figure} 

    \end{enumerate}

    \item \textbf{Police Concentration Areas in Baltimore}
    
    This subsection focuses on how police resources are distributed across the neighborhoods over the course of the simulation. To be more precise, we looked at the number of scenarios in which each of the top 10 most frequently over-policed neighborhoods were ranked as top-3. 
    \begin{enumerate}[label=\textbf{Finding~\arabic{enumi}.\arabic{enumii}}]
        \item {\textbf{Assigning Officers Based on Aggravated Assault Over-policed more Black Neighborhoods than White Neighborhoods}}
        
        Over-policing is defined as having a higher police share compared to crime share. Looking at the top 10 most frequently over-policed neighborhoods by each algorithm when aggravated assault records were used, we observed that other than \textbf{Downtown}, which is the most frequently over-policed neighborhood in the scenarios, all other top 10 frequently over-policed neighborhoods were Black, such as \textbf{Blair-Edison} and \textbf{Sandtown-Winchester}. (\cref{Top10_Overpoliced_agg}).
        \begin{figure}[H]
            \begin{flushright}
                \begin{minipage}{0.85\linewidth}
                    \includegraphics[width=\linewidth, height=.6\textwidth]{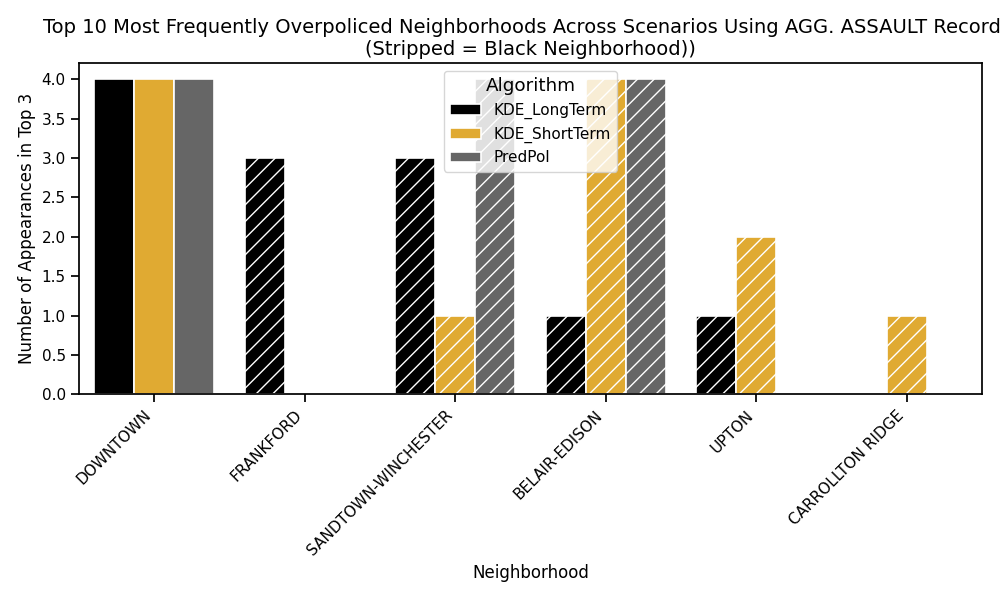}
                    \caption{Top 10 neighborhoods across scenarios that appeared in the list of top 3 over-policed neighborhoods in each individual scenario when applying aggravated assault records by each algorithm.}
                    \label{Top10_Overpoliced_agg}
                \end{minipage}
            \end{flushright}
        \end{figure}
        
        \item {\textbf{Assigning Officers Based on all Crime Records by PredPol Over-policed Mostly White Neighborhoods, Such as Mount Vernon and Canton, Unlike for the Other two Models.}}\\
        Looking at the top 10 most frequently over-policed neighborhoods by each algorithm when all crime records were used, we saw that other than \textbf{Sandtown-Winchester and Cherry Hill}, all other top 10 frequently over-policed neighborhoods by \textbf{PredPol} were White, such as \textbf{Mount Vernon}, \textbf{Canton}, and \textbf{Brooklyn}.  (\cref{Top10_Overpoliced_total}). While among \textbf{short-term KDE}'s over-policed neighborhoods, only two out of 6 (\textbf{Paterson Park and Inner Harbor}) and among \textbf{long-term KDE}'s one out of four (\textbf{Brooklyn}) were white.
        \begin{figure}[H]
            \begin{flushright}
                \begin{minipage}{0.85\linewidth}
                    \includegraphics[width=\linewidth, height=.6\textwidth]{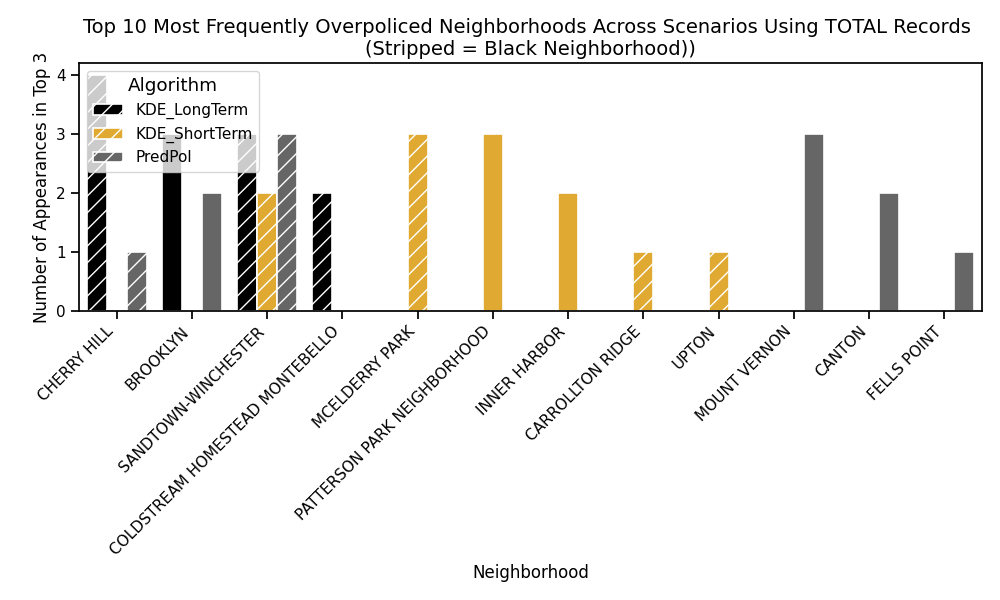}
                    \caption{Top 10 neighborhoods across scenarios that are appearing in the list of top 3 over-policed neighborhoods in each individual scenario, when applying total crime records, by each algorithm.}
                    \label{Top10_Overpoliced_total}
                \end{minipage}
            \end{flushright}
        \end{figure}
        
    \end{enumerate}
\end{enumerate}
\section{Discussion}

\label{sec:Discussion}

\jf{We performed simulations  in a range of scenarios comparing different policing systems, using a Baltimore crime dataset. Based on our simulations and analyses we drew the following overall conclusions.}
\begin{enumerate}
    \item \textbf{Fairness and Accuracy Comparison}
    
    By comparing simulation results of predictive policing and hot spots policing representatives, 
    \jf{we} discovered that \jf{predictive policing via} PredPol was generally 
    \jf{more} accurate and had higher racial and neighborhood-level fairness than \jf{hot spots policing via} short-term and long-term KDE across most scenarios.
    
    \jf{Long-term KDE, where we widened KDE's data input to use the same data PredPol received,}  
    improved the accuracy and neighborhood-level fairness \jf{over short-term KDE} but worsened the racial fairness. The improvement in accuracy did not overtake PredPol's level of accuracy but it almost matched its neighborhood-level fairness. 
    
    The theory of hot spots policing 
    \jf{claims that the} recent crimes' distribution can approximate \jf{their} near future distribution. The recent crimes are often defined as crime history within a month or a year of the prediction date~\cite{halford2024hotspot}. In our experiment, both Long-term KDE and PredPol 
    \jf{received} the crime history from a year before the start day of the simulation. The simulation's starting prediction date is January first, 2019, and the crime history is all the crimes after January first, 2018. If the crime history was more than a year ago, we 
    \jf{expect} that the 
    \jf{accuracy at the starting date} would drop, but overall accuracy might improve because 
    \jf{our experiments found that for KDE,}   
    a longer crime history slows down the effect of neighborhood-level bias, which slows down the accuracy drop.
    
    \jf{Our results showed that h}igher accuracy does not guarantee a racially fairer model. Both fair and unfair outcomes were observed across the fairness spectrum for high-accuracy scenarios (\cref{accuracy_vs_Racefairness}). However, neighborhood-level fairness seemed to correlate with accuracy (\cref{accuracy_vs_Neifairness}). These results extend Mohler et al.’s work~\cite{mohler2018penalized} on accuracy–fairness trade-offs by providing evidence from extended simulations on real crime data. Unlike Mohelr's observation that fairness came at a cost for accuracy, it was demonstrated that some scenarios were both fairer and more accurate than others.  This can also be  inferred that the give-or-take between accuracy and fairness depends on the fairness metric and also the demographic geography of the location the system is being customized for (\cref{accuracy_vs_Neifairness}, and~\cref{accuracy_vs_Racefairness}).
    
    \item \textbf{Temporal Feedback Loop Effects}
    
    In most of the scenarios with all three systems, we saw a trend of bias amplification unless there was a change in the pattern of data causing the police focus to change from certain neighborhoods to others during the simulation. In scenarios where bias was not worsening, the race that got the highest focus at the start of the simulation contradicted the one at the end, as 
    \jf{shown in} in \cref{AlgorithmCompare_WorseningOrImproving_racial_fairnessGap_slope_avgPolShare} and~\cref{AlgorithmCompare_WorseningOrImproving_racial_fairnessGap_slope_PCR}. 
    \jf{Note} that the negative slopes of fairness gaps were the scenarios with improvement in regard to that metric. For these scenarios bias amplification might have happened later if the simulation was continued for a longer period unless reported crime distribution changed drastically again. Therefore, using real crime data, we saw that bias amplification was not constant and it depended on the changes in the newly discovered and reported crime distribution. 
    For 
    \jf{n}eighborhood\jf{-}level fairness, we saw a 
    \jf{continual} bias rise in most of the scenarios, especially for \jf{P}red\jf{P}ol and short-term KDE (~\cref{AlgorithmCompare_WorseningOrImproving_NeighborhoodLevel_fairnessGap_pcr}, and~\cref{AlgorithmCompare_WorseningOrImproving_NeighborhoodLevel_fairnessGap_pcr}). For racial fairness, a lower number of scenarios experienced the bias amplification compared to neighborhood-level, especially for the metric related to average police share (\cref{AlgorithmCompare_WorseningOrImproving_racial_fairnessGap_slope_avgPolShare}). However, it was noted that all those with bias improvement had a contradiction between the race the police focus was on and the race the focus was trending toward.
    
    \jf{It is important to consider that} bias amplification was defined as the falling trend of a system's equality of treatment. In all of the scenarios in which we saw improvement in equality, it was also observed that the race that was receiving a higher police share at the beginning of the simulation and the race the trend was toward were different. Therefore, a longer simulation duration is needed in order to make sure there 
    \jf{is not} a point of racial equality in the future, after which a bias amplification toward the trending race occurs. We did not extend the simulation further, as longer durations introduce a shift in the underlying crime distribution that interacts with the feedback effect, making it difficult to attribute the changes in the fairness to a single mechanism. Unlike Ensign's assumptive theoretical study that showed a constant bias amplification when assigning an officer between two neighborhoods~\cite{ensign2018runaway}, in a more real-world-like situation the amplification is not constant; it could temporarily subside and then intensify again.

    Although PredPol was generally fairer and more accurate on average in most of the scenarios, its speed of bias amplification was higher for most of the metrics in most of the scenarios, while the speed of bias amplification for long-term KDE was 
    \jf{substantially} slower compared to the other two in most of the scenarios. This was further 
    \jf{evidence} that different models respond differently to feedback loops.

    \item \textbf{Crime Type Effect}
    
    In the context of Baltimore, 
    \jf{u}sing different data from all crime records flipped the police concentration from one demographic to another for PredPol and, in some scenarios, for short-term and long-term KDE. 
    \jf{This result showed} that the direction of racial bias could be affected based \jf{not only} on the predictive policing system, 
    \jf{but also} the data that \jf{it} was applied \jf{to}.
    
    \jf{We also observed that} the average accuracy and average neighborhood-level equality based on police-to-crime ratio dropped for long-term KDE and PredPol. 
    \jf{This suggests} that systems with more focus on long-term effects of crime on future crime prediction might become less accurate when the daily records are sparse
    \jf{. Alternatively, it could be} because that data record had a higher change in distribution in a short period of time. The sparseness of crime data and its effect on different systems could be studied further \jf{in future work}.
    
    \item \textbf{Data-Driven vs Model-Driven Bias}
    
    \jf{Our analyses found that b}oth data and model matter: \jf{the} 
    \jf{s}ame data fed into different models (KDE vs. PredPol) produced different fairness outcomes as previously shown by Chapman et al.~\cite{chapman2022data}. The same model fed different data (aggravated assault vs. total crime) can 
    \jf{affect the speed of bias development, or even flip the direction of bias.}
    
    \item \textbf{Baltimore-Specific Insights}
    
    Black neighborhoods received more police on average in most simulations—but this varied with crime type and algorithm. In most scenarios the trend was toward assigning a higher average share of officers in general and also a higher average share of officers per share of crime to White neighborhoods when all crime records were used. \jf{This finding countered the widely-held assumption that predictive policing is biased specifically against Black neighborhoods, as found by Lum and Isaac in Oakland data~\cite{lum2016predict}.} These patterns are not solely explainable by neighborhood count (Baltimore has more Black than White neighborhoods), suggesting model behavior plays a larger role than geography alone.
    
    Downtown, Sandtown-Winchester, and Blair-Edison were repeatedly among the top neighborhoods being over-policed when aggravated assault records were used. This consistency across models suggests that some neighborhoods were structurally favored/targeted, regardless of the predictive algorithm.
    
    Looking at the over-policed neighborhoods when all crime records were used, however, we see that models were not as consistent, \jf{which was} another demonstration of how data and algorithms both affect the results. The only neighborhood 
    \jf{the algorithms} 
    \jf{commonly over-policed} when total crime records were used was Sandtown-Winchester, a Black neighborhood.
    
    These Baltimore predictive policing simulation results were consistent with 
    \jf{previous studies on other cities}, whether by real experimental studies~\cite{mohler2015randomized} or by simulation studies~\cite{lum2016predict}, \jf{in that police concentration tendencies occurred in these models due to feedback loops.}
    
    Based on the duration of the simulation, these systems might change rank in terms of each bias metric or even average accuracy. A short simulation duration might indicate PredPol as causing the most uniform police distribution (based on police Gini coefficient), followed by long-term KDE and then short-term KDE, while a longer simulation duration might cause a rank swap between long-term KDE and PredPol. These variables could be studied in future system-comparative works. 

    For Baltimore City our recommendations are as follows: 
    \begin{itemize}
        \item Although predictive policing remained fairer and more accurate than hotspot policing, it had a higher speed of bias amplification in most scenarios. Hence, we advise the city to be aware of the long-term tendencies of any predictive policing system they might use. 
        \item \jf{M}ake sure to perform pre-real-world implementation evaluation studies like 
        \jf{the one we performed here}. These types of studies could 
        \jf{highlight bias issues arising from the combination of a particular algorithm and data,} and would make the decision-making better informed.
        \item In resource distribution using the predictive policing model, based on aggravated assault records, 
        \jf{authorities} should be mindful of assigning too many officers to Downtown, Blair-Edison, and Sandtown-Winchester \jf{neighborhoods}. 
        \jf{We advise that} when applying all crime records to the predictive policing model, Sandtown-Winchester, Brooklyn, Mount Vernon, and Canton could 
        \jf{incorrectly appear} to have higher crime rates \jf{due to feedback loop phenomena.} 
        \jf{Similarly,} the city can prepare a list of neighborhoods that might be underrated and under-policed and in need of more attention. 
        
    \end{itemize}

\end{enumerate}
\section{Limitations}
Limitations of the work \jf{include:} 
    \begin{itemize}
        \item We assume the crimes in the database are all \jf{of} the crimes that have actually happened, which is untrue. These records could be the result of an already biased system by having less crime reported or discovered in some areas. 
        \item 
        \jf{We} used real crime data and filtered it by a fixed report probability for all neighborhoods. 
        \jf{It would be more realistic} if different crimes were filtered based on their report probability rather than an average report probability for all. 
        \jf{Modeling d}ifferent report probabilities of different neighborhoods would \jf{also} affect the results. 
        \item We did not account for how the presence of police in a neighborhood could affect the crime rate. 
        \jf{Currently,} we 
        \jf{expect that by} running the simulation for longer periods, if 
        the distribution of those filtered reported crimes changes by suddenly having several days of high crime rates in neighborhoods 
        \jf{with different demographic majorities}, 
        these systems could \jf{potentially even} flip \jf{their} bias direction tendency.
        \item In this work, we consider\jf{ed} a detection formula that implicitly assumes crime detection is determined exclusively by the number of deployed officers. Nevertheless, in real-world deployments, detection rates may also be affected by the model-estimated level of crime risk itself. Officers entering a prediction area typically behave with elevated vigilance, which can inform how they evaluate behavior and distribute attention~\cite{ferguson2012predictive}. As a result, detection probability may be determined not only by officer count but also by the broader sociotechnical context induced by the predictive mechanism.
    \end{itemize} 

\section{Conclusion}

We developed an agent-based modeling simulation to investigate whether the use of hot-spot policing (short-term and long-term KDE) and predictive policing (PredPol) lead to feedback loops and hence racial bias in Baltimore
\jf{. We} defined accuracy and fairness metrics to evaluate and compare the systems' 
behavior \jf{in this locality}. In doing so, we extended the body of literature by providing a framework consisting of fairness and accuracy metrics to compare and evaluate different systems' long-term localized tendencies and provided insights into the resource distribution tendencies of each model, recounting the neighborhoods that might be over- or underrated as crime hotspots by looking at which neighborhoods will be over-policed or under-policed during a 300 day simulation period.  Our results on 2019-2020 data from Baltimore City showed that predictive policing and hot-spot policing are indeed subject to feedback loops and racial and neighborhood-level bias. Although the speed of bias amplification was higher for predictive policing, it is the fairer and more accurate system on average compared to the hot-spot policing systems. 

The findings from Baltimore show that, contrary to the generally believed bias trends, in these 300-day simulation scenarios, when using all crime records, the police concentration trends were mostly toward White neighborhoods. Downtown Baltimore is a mostly White-dominant area, which appeared as the neighborhood receiving the highest police shares (with a rising trend) in most of the simulation scenarios. \jf{This shows that fairness issues in predictive policing algorithms have impacts beyond the Black community, which could motivate a broader set of stakeholders to work together to find solutions that benefit the entire community.} Our results emphasize the need for localized fairness assessments and cautious use of predictive policing tools in sensitive environments. 


\section{Acknowledgments}
We extend our gratitude to the undergraduate and graduate students who previously contributed to this project, laying the groundwork for the research presented here. Although their specific contributions 
\jf{did not appear} in this paper, their efforts were valuable in preparing for this study. 
We would like to thank the graduate student researchers Sambhaw Sharma, Ashwathy Samivel Sureshkumar, Harish Ramamoorthy, Akarshika Singhal, Vamshi Krishna Yenmangandla, Pranvat Singh, and Dharmil Shah, whose coding contributions aided the preliminary exploratory analysis, and the undergraduate student researchers Aminat Alabi and Shaniah Reece for the literature review and critical analysis of the local context.


\textbf{Funding}

This material is based upon work supported by the National Science Foundation under Grant No.’s
IS1927486; IIS2046381. Any opinions, findings, and conclusions or recommendations expressed in
this material are those of the author(s) and do not necessarily reflect the views of the National
Science Foundation.

\textbf{Data Availability}

The datasets used in this study were originally retrieved from the Baltimore City Open Data Portal. As they are no longer available at the original links, archived copies along with the code are provided at: \url{https://github.com/saminsemsar/Data_Analysis_Portfolio/tree/main/PredictivePolicing}.


\bibliographystyle{plain}
\bibliography{references}

@misc{BaltimoreSun,
  author       = {Prudente, Tim},
  title        = {High-tech policing coming to Baltimore},
  year         = {2018},
  month        = {feb},
  day          = {1},
  publisher    = {Baltimore Sun Digital Edition}
}

@misc{FoxBaltimore,
  author       = {Zumer, Bryna},
  title        = {Baltimore Police Department to launch predictive policing strategy},
  year         = {2018},
  note         = {Accessed January 9, 2025},
  howpublished = {\url{https://foxbaltimore.com/news/local/baltimore-police-to-launch-predictive-policing-strategy}},
  publisher    = {Fox Baltimore}
}

@article{zubair2025crime,
  title={Crime Hotspot Prediction Using Deep Graph Convolutional Networks},
  author={Zubair, Tehreem and Fatima, Syeda Kisaa and Ahmed, Noman and Khan, Asifullah},
  journal={arXiv preprint arXiv:2506.13116},
  year={2025}
}

@article{mohler2011self,
  title={Self-exciting point process modeling of crime},
  author={Mohler, George O and Short, Martin B and Brantingham, P Jeffrey and Schoenberg, Frederic Paik and Tita, George E},
  journal={Journal of the american statistical association},
  volume={106},
  number={493},
  pages={100--108},
  year={2011},
  publisher={Taylor \& Francis}
}

@article{braga2019hot,
  title={Hot spots policing of small geographic areas effects on crime},
  author={Braga, Anthony A and Turchan, Brandon and Papachristos, Andrew V and Hureau, David M},
  journal={Campbell systematic reviews},
  volume={15},
  number={3},
  pages={e1046},
  year={2019},
  publisher={Wiley Online Library}
}

@article{bowers2004prospective,
  title={Prospective hot-spotting: the future of crime mapping?},
  author={Bowers, Kate J and Johnson, Shane D and Pease, Ken},
  journal={British journal of criminology},
  volume={44},
  number={5},
  pages={641--658},
  year={2004},
  publisher={Oxford University Press}
}

@book{perry2013predictive,
  title={Predictive policing: The role of crime forecasting in law enforcement operations},
  author={Perry, Walt L and McInnis, B. and Price, Carter C. and
Smith, Susan C.  and Hollywood, John S. },
  year={2013},
  publisher={Rand Corporation}
}

@article{lum2016predict,
  title={To predict and serve?},
  author={Lum, Kristian and Isaac, William},
  journal={Significance},
  volume={13},
  number={5},
  pages={14--19},
  year={2016},
  publisher={Oxford University Press}
}

@misc{ACLU2010Settlement,
  author       = {{American Civil Liberties Union}},
  title        = {Plaintiffs Win Justice in Illegal Arrests Lawsuit Settlement with Baltimore City Police},
  year         = {2010},
  howpublished = {\url{https://www.aclu.org/press-releases/plaintiffs-win-justice-illegal-arrests-lawsuit-settlement-baltimore-city-police}},
  note         = {Accessed: 2025-09-08}
}

@misc{ACLU2012NonCompliance,
  author       = {{American Civil Liberties Union}},
  title        = {ACLU Condemns Baltimore Police Department for Failing to Comply with Settlement Agreement in Illegal Arrests Case},
  year         = {2012},
  howpublished = {\url{https://www.aclu.org/press-releases/aclu-condemns-baltimore-police-department-failing-comply-settlement-agreement-illegal}},
  note         = {Accessed: 2025-09-08}
}

@misc{DOJ2016Findings,
  author       = {{United States Department of Justice}},
  title        = {Investigation of the Baltimore City Police Department},
  year         = {2016},
  month        = {August},
  publisher    = {U.S. Department of Justice, Civil Rights Division},
  howpublished = {\url{https://www.justice.gov/archives/opa/file/883366/dl?inline}},
  note         = {Accessed: 2025-09-08}
}

@book{brown2021black,
  title={The black butterfly: The harmful politics of race and space in America},
  author={Brown, Lawrence T},
  year={2021},
  publisher={JHU Press}
}

@book{pietila2010not,
  title={Not in my neighborhood: How bigotry shaped a great American city},
  author={Pietila, Antero},
  year={2010},
  publisher={Bloomsbury Publishing USA}
}

@misc{Time2015FreddieGray,
  author       = {Time Magazine},
  title        = {Baltimore Protests Turn Violent After Freddie Gray’s Funeral},
  year         = {2015},
  month        = {April},
  howpublished = {\url{https://time.com/3837454/baltimore-looting-clashes-freddie-gray-police-protesters}},
  note         = {Accessed: 2025-09-08}
}

@misc{VanityFair2015FreddieGray,
  author       = { Kia Makarechi},
  title        = {The Clock Didn’t Start with the Riots: Baltimore and Freddie Gray},
  year         = {2015},
  month        = {April},
  publisher    = {Vanity Fair},
  howpublished = {\url{https://www.vanityfair.com/news/2015/04/baltimore-riots-freddie-gray}},
  note         = {Accessed: 2025-09-08}
}

@article{hanlon2007fate,
  title={The fate of inner suburbs: Evidence from metropolitan Baltimore},
  author={Hanlon, Bernadette and Vicino, Thomas J},
  journal={Urban Geography},
  volume={28},
  number={3},
  pages={249--275},
  year={2007},
  publisher={Taylor \& Francis}
}

@book{alexander2012new,
  author    = {Alexander, Michelle},
  title     = {The New Jim Crow: Mass Incarceration in the Age of Colorblindness},
  publisher = {The New Press},
  year      = {2012},
  address   = {New York},
  edition   = {Revised edition}
}

@techreport{NoBoundaries2016Report,
  author       = {{No Boundaries Coalition}},
  title        = {Over-Policed, Yet Underserved: The People’s Findings Regarding Police Encounters and Accountability in Central West Baltimore},
  year         = {2016},
  institution  = {No Boundaries Coalition},
  url          = {https://www.noboundariescoalition.com/wp-content/uploads/2016/03/No-Boundaries-Layout-Web-1.pdf},
  note         = {Accessed: 2025-11-10}
}

@incollection{densley2021over,
  title={Over-policed and under-protected: Police violence as a symptom and cause of urban violence in America's Black communities},
  author={Densley, James A},
  booktitle={Public health, mental health, and mass atrocity prevention},
  pages={71--88},
  year={2021},
  publisher={Routledge}
}

@misc{cbs2015freddiegray,
  author       = {{CBS News}},
  title        = {Violent crime rate spikes in Baltimore after Freddie Gray's death in police custody},
  howpublished = {\url{https://www.cbsnews.com/news/violent-crime-rate-spikes-baltimore-freddie-gray-death-police-custody-2015/}},
  year         = {2015},
  note         = {Accessed: 2025-11-12}
}

@misc{fivethirtyeight2015baltimore,
  author       = {Koerth-Baker, Maggie and Bronner, Laura},
  title        = {Charts: Baltimore Crime Before And After Freddie Gray's Funeral},
  howpublished = {\url{https://www.fivethirtyeight.com/features/charts-baltimore-crime-before-and-after-freddie-grays-funeral/}},
  year         = {2015},
  note         = {FiveThirtyEight, Accessed: 2025-11-12}
}

@misc{pew2015baltimore,
  author       = {{Pew Research Center}},
  title        = {Multiple Causes Seen for Baltimore Unrest},
  howpublished = {\url{https://www.pewresearch.org/politics/2015/05/04/multiple-causes-seen-for-baltimore-unrest/}},
  year         = {2015},
  note         = {Accessed: 2025-11-12}
}

@misc{latimes2015baltimore,
  author       = {{Los Angeles Times}},
  title        = {CVS pharmacy emerges as symbolic flashpoint of Baltimore riot},
  howpublished = {\url{https://www.latimes.com/nation/la-na-cvs-pharmacy-baltimore-riots-20150428-story.html}},
  year         = {2015},
  note         = {Accessed: 2025-11-12}
}

@misc{cbc2015baltimore,
  author       = {{CBC News}},
  title        = {Baltimore riots prompt state of emergency after Freddie Gray's funeral},
  howpublished = {\url{https://www.cbc.ca/news/world/baltimore-riots-prompt-state-of-emergency-after-freddie-gray-funeral-1.3051048}},
  year         = {2015},
  note         = {Accessed: 2025-11-12}
}

@misc{cnn2015baltimore,
  author       = {{CNN}},
  title        = {Baltimore drug market disrupted after Freddie Gray's death},
  howpublished = {\url{https://www.cnn.com/2015/06/25/politics/baltimore-drug-market-freddie-gray}},
  year         = {2015},
  note         = {Accessed: 2025-11-12}
}

@article{Guardian2016FreddieGrayTimeline,
  author  = {{The Guardian}},
  title   = {Freddie Gray: From Arrest to Protest — A Timeline},
  journal = {The Guardian},
  year    = {2016},
  month   = {April},
  howpublished = {\url{https://www.theguardian.com/us-news/2016/apr/27/baltimore-freddie-gray-arrest-protest-timeline}},
  note    = {Accessed: 2025-11-10}
}

@misc{BALT,
  author    = {{Baltimore Action Legal Team}},
  title     = {About Us},
  year      = {2024},
  howpublished = {\url{https://www.baltimoreactionlegal.org/aboutus}},
  note      = {Accessed: 2025-11-10}
}

@misc{NoBoundaries,
  author    = {{No Boundaries Coalition}},
  title     = {About Us},
  year      = {2024},
  howpublished = {\url{https://www.noboundariescoalition.com/about-us/}},
  note      = {Accessed: 2025-11-10}
}

@misc{LBS,
  author    = {{Leaders of a Beautiful Struggle}},
  title     = {About},
  year      = {2024},
  howpublished = {\url{https://lbsbaltimore.com/about/}},
  note      = {Accessed: 2025-11-10}
}

@article{parzen1962estimation,
  title={On estimation of a probability density function and mode},
  author={Parzen, Emanuel},
  journal={The annals of mathematical statistics},
  volume={33},
  number={3},
  pages={1065--1076},
  year={1962},
  publisher={JSTOR}
}

@article{Rosenblatt1956RemarksOS,
  title={Remarks on Some Nonparametric Estimates of a Density Function},
  author={Murray Rosenblatt},
  journal={Annals of Mathematical Statistics},
  year={1956},
  volume={27},
  pages={832-837},
  url={https://api.semanticscholar.org/CorpusID:16643156}
}

@article{chainey2008utility,
  title={The utility of hotspot mapping for predicting spatial patterns of crime},
  author={Chainey, Spencer and Tompson, Lisa and Uhlig, Sebastian},
  journal={Security journal},
  volume={21},
  number={1},
  pages={4--28},
  year={2008},
  publisher={Springer}
}

@article{hu2018spatio,
  title={A spatio-temporal kernel density estimation framework for predictive crime hotspot mapping and evaluation},
  author={Hu, Yujie and Wang, Fahui and Guin, Cecile and Zhu, Haojie},
  journal={Applied geography},
  volume={99},
  pages={89--97},
  year={2018},
  publisher={Elsevier}
}

@article{mohler2014marked,
  title={Marked point process hotspot maps for homicide and gun crime prediction in Chicago},
  author={Mohler, George},
  journal={International Journal of Forecasting},
  volume={30},
  number={3},
  pages={491--497},
  year={2014},
  publisher={Elsevier}
}

@misc{uspatent8949164,
  author       = {Mohler, George O. and Short, Martin B. and Malinowski, Sean and Johnson, Mark and Tita, George},
  title        = {Systems and methods for predictive policing: US Patent US8949164B1},
  howpublished = {\url{https://patents.google.com/patent/US8949164B1/en}},
  year         = {2015},
  note         = {Accessed: 2025-11-12}
}

@article{vivek2023spatio,
  title={Spatio-temporal crime analysis and forecasting on twitter data using machine learning algorithms},
  author={Vivek, Meghashyam and Prathap, Boppuru Rudra},
  journal={SN Computer Science},
  volume={4},
  number={4},
  pages={383},
  year={2023},
  publisher={Springer}
}

@article{tam2023multimodal,
  title={Multimodal deep learning crime prediction using tweets},
  author={Tam, Sakirin and Tanri{\"o}ver, {\"O}mer {\"O}zg{\"u}r},
  journal={IEEE Access},
  volume={11},
  pages={93204--93214},
  year={2023},
  publisher={IEEE}
}

@article{barbosa2015exploiting,
  title={Exploiting spatio-temporal patterns using partial-state reinforcement learning in a synthetically augmented environment},
  author={Barbosa, Salvador E and Petty, Mikel D},
  journal={Progress in Artificial Intelligence},
  volume={3},
  pages={55--71},
  year={2015},
  publisher={Springer}
}

@inproceedings{joe2022reinforcement,
  title={Reinforcement learning approach to solve dynamic bi-objective police patrol dispatching and rescheduling problem},
  author={Joe, Waldy and Lau, Hoong Chuin and Pan, Jonathan},
  booktitle={Proceedings of the International Conference on Automated Planning and Scheduling},
  volume={32},
  pages={453--461},
  year={2022}
}

@inproceedings{chen2023risk,
  title={A risk-aware multi-objective patrolling route optimization method using reinforcement learning},
  author={Chen, Haowen and Wu, Yifan and Wang, Weikun and Zheng, Zengwei and Ma, Jianhua and Zhou, Binbin},
  booktitle={2023 IEEE 29th International Conference on Parallel and Distributed Systems (ICPADS)},
  pages={1637--1644},
  year={2023},
  organization={IEEE}
}

@article{chen2023optimizing,
  title={Optimizing Patrolling Route with a Risk-Aware Reinforcement Learning Model},
  author={Chen, Haowen and Wu, Yifan and Wang, Weikun and Zheng, Zengwei and Ma, Jianhua and Zhou, Binbin},
  year ={2023},
  journal={Available at SSRN 4752931}
}

@article{repasky2024multi,
  title={Multi-Agent Reinforcement Learning for Joint Police Patrol and Dispatch},
  author={Repasky, Matthew and Wang, He and Xie, Yao},
  journal={arXiv preprint arXiv:2409.02246},
  year={2024}
}

@article{joe2023learning,
  title={Learning to send reinforcements: coordinating multi-agent dynamic police patrol dispatching and rescheduling via reinforcement learning},
  author={Joe, Waldy and Lau, Hoong Chuin},
  year={2023},
  publisher={AAAI Press}
}

@article{brantingham2017logic,
  title={The logic of data bias and its impact on place-based predictive policing},
  author={Brantingham, P Jeffrey},
  journal={Ohio St. J. Crim. L.},
  volume={15},
  pages={473},
  year={2017},
  publisher={HeinOnline}
}

@inproceedings{ensign2018runaway,
  title={Runaway feedback loops in predictive policing},
  author={Ensign, Danielle and Friedler, Sorelle A and Neville, Scott and Scheidegger, Carlos and Venkatasubramanian, Suresh},
  booktitle={Conference on fairness, accountability and transparency},
  pages={160--171},
  year={2018},
  organization={PMLR}
}

@article{brantingham2018does,
  title={Does predictive policing lead to biased arrests? Results from a randomized controlled trial},
  author={Brantingham, P Jeffrey and Valasik, Matthew and Mohler, George O},
  journal={Statistics and public policy},
  volume={5},
  number={1},
  pages={1--6},
  year={2018},
  publisher={Taylor \& Francis}
}

@inproceedings{mohler2018penalized,
  title={A penalized likelihood method for balancing accuracy and fairness in predictive policing},
  author={Mohler, George and Raje, Rajeev and Carter, Jeremy and Valasik, Matthew and Brantingham, Jeffrey},
  booktitle={2018 IEEE international conference on systems, man, and cybernetics (SMC)},
  pages={2454--2459},
  year={2018},
  organization={IEEE}
}

@inproceedings{akpinar2021effect,
  title={The effect of differential victim crime reporting on predictive policing systems},
  author={Akpinar, Nil-Jana and De-Arteaga, Maria and Chouldechova, Alexandra},
  booktitle={Proceedings of the 2021 ACM conference on fairness, accountability, and transparency},
  pages={838--849},
  year={2021}
}

@inproceedings{chapman2022data,
  title={A Data-driven analysis of the interplay between Criminological theory and predictive policing algorithms},
  author={Chapman, Adriane and Grylls, Philip and Ugwudike, Pamela and Gammack, David and Ayling, Jacqui},
  booktitle={Proceedings of the 2022 ACM Conference on Fairness, Accountability, and Transparency},
  pages={36--45},
  year={2022}
}

@inproceedings{mashiat2023counterfactually,
  title={Counterfactually fair dynamic assignment: A case study on policing},
  author={Mashiat, Tasfia and Gitiaux, Xavier and Rangwala, Huzefa and Das, Sanmay},
  booktitle={Proceedings of the 2023 International Conference on Autonomous Agents and Multiagent Systems},
  pages={2526--2528},
  year={2023}
}

@article{griffard2019bias,
  title={A Bias-Free Predictive Policing Tool: An Evaluation of the NYPD's Patternizr},
  author={Griffard, Molly},
  journal={Fordham Urb. LJ},
  volume={47},
  pages={43},
  year={2019},
  publisher={HeinOnline}
}

@article{lagioia2023algorithmic,
  title={Algorithmic fairness through group parities? The case of COMPAS-SAPMOC},
  author={Lagioia, Francesca and Rovatti, Riccardo and Sartor, Giovanni},
  journal={AI \& society},
  volume={38},
  number={2},
  pages={459--478},
  year={2023},
  publisher={Springer}
}

@article{wang2019empirical,
  title={An empirical study on learning fairness metrics for compas data with human supervision},
  author={Wang, Hanchen and Grgic-Hlaca, Nina and Lahoti, Preethi and Gummadi, Krishna P and Weller, Adrian},
  journal={arXiv preprint arXiv:1910.10255},
  year={2019}
}

@article{dressel2018accuracy,
  title={The accuracy, fairness, and limits of predicting recidivism},
  author={Dressel, Julia and Farid, Hany},
  journal={Science advances},
  volume={4},
  number={1},
  pages={eaao5580},
  year={2018},
  publisher={American Association for the Advancement of Science}
}

@article{mohler2015randomized,
  title={Randomized controlled field trials of predictive policing},
  author={Mohler, George O and Short, Martin B and Malinowski, Sean and Johnson, Mark and Tita, George E and Bertozzi, Andrea L and Brantingham, P Jeffrey},
  journal={Journal of the American statistical association},
  volume={110},
  number={512},
  pages={1399--1411},
  year={2015},
  publisher={Taylor \& Francis}
}

@techreport{helms2020memphis,
  author      = {Helms, James Max and Madden, Angela},
  title       = {Assessment of Data-Driven Deployment by the Memphis Police Department},
  institution = {Public Safety Institute, University of Memphis},
  year        = {2020},
  note        = {Fall 2020 Report},
  url         = {https://www.memphis.edu/psi/pdfs/2021-cas-312-data-driven-deployment-assessment.pdf},
  type        = {Technical Report}
}

@article{rajipredictive,
  title={Predictive Policing: The Role of AI in Crime Prevention},
  author={Raji, Ibrahim and Sholademi, Damilola Bartholomew}
}

@article{mandalapu2023crime,
  title={Crime prediction using machine learning and deep learning: A systematic review and future directions},
  author={Mandalapu, Varun and Elluri, Lavanya and Vyas, Piyush and Roy, Nirmalya},
  journal={Ieee Access},
  volume={11},
  pages={60153--60170},
  year={2023},
  publisher={IEEE}
}

@misc{baltimore_crime_data,
  author       = {{Baltimore Police Department}},
  title        = {{Part 1 Crime Data}},
  year         = {n.d.},
  note         = {Accessed: 2023-03-11},
  url          = {https://data.baltimorecity.gov/datasets/baltimore::part-1-crime-data/explore?location=18.619939%2C-38.355639%2C4.62&showTable=true},
  howpublished = {\url{https://data.baltimorecity.gov/datasets/baltimore::part-1-crime-data}}
}

@misc{baltimore_neighborhood_data,
  author       = {{City of Baltimore}},
  title        = {{Neighborhood Demographic and Spatial Data}},
  year         = {n.d.},
  note         = {Accessed: 2022-08-23},
  url          = {https://data.baltimorecity.gov/datasets/neighborhood-1/explore?location=39.284818%2C-76.620500%2C11.47&showTable=true},
  howpublished = {\url{https://data.baltimorecity.gov/datasets/neighborhood-1}}
}

@misc{baltimore_neighborhood_boundaries,
  author       = {{City of Baltimore}},
  title        = {{Neighborhood Boundary KML File}},
  year         = {n.d.},
  note         = {Accessed: 2023-03-15},
  url          = {https://data.baltimorecity.gov/datasets/baltimore::neighborhood-1/explore?location=39.284823%2C-76.620500%2C11.51},
  howpublished = {\url{https://data.baltimorecity.gov/datasets/baltimore::neighborhood-1}}
}

@techreport{bjs_crime_2019,
  author      = {Morgan, Rachel E. and Truman, Jennifer L.},
  title       = {Criminal Victimization, 2019},
  institution = {Bureau of Justice Statistics, U.S. Department of Justice},
  year        = {2020},
  number      = {NCJ 255113},
  url         = {https://bjs.ojp.gov/content/pub/pdf/cv19.pdf},
  note        = {Accessed: June 13, 2025}
}

@article{scikit_learn,
  author  = {Fabian Pedregosa and Ga{{\"e}}l Varoquaux and Alexandre Gramfort and Vincent Michel and Bertrand Thirion and Olivier Grisel and Mathieu Blondel and Peter Prettenhofer and Ron Weiss and Vincent Dubourg and Jake Vanderplas and Alexandre Passos and David Cournapeau and Matthieu Brucher and Matthieu Perrot and {{\'E}}douard Duchesnay},
  title   = {Scikit-learn: Machine Learning in Python},
  journal = {Journal of Machine Learning Research},
  year    = {2011},
  volume  = {12},
  number  = {85},
  pages   = {2825--2830},
  url     = {http://jmlr.org/papers/v12/pedregosa11a.html}
}

@inproceedings{dwork2012fairness,
  title={Fairness through awareness},
  author={Dwork, Cynthia and Hardt, Moritz and Pitassi, Toniann and Reingold, Omer and Zemel, Richard},
  booktitle={Proceedings of the 3rd innovations in theoretical computer science conference},
  pages={214--226},
  year={2012}
}

@inproceedings{friedler2019comparative,
  title={A comparative study of fairness-enhancing interventions in machine learning},
  author={Friedler, Sorelle A and Scheidegger, Carlos and Venkatasubramanian, Suresh and Choudhary, Sonam and Hamilton, Evan P and Roth, Derek},
  booktitle={Proceedings of the conference on fairness, accountability, and transparency},
  pages={329--338},
  year={2019}
}

@article{abeles2020gini,
  title={The Gini coefficient as a useful measure of malaria inequality among populations},
  author={Abeles, Jonathan and Conway, David J},
  journal={Malaria journal},
  volume={19},
  number={1},
  pages={444},
  year={2020},
  publisher={Springer}
}

@article{thomas2003measuring,
  title={Measuring education inequality: Gini coefficients of education for 140 countries, 1960-2000},
  author={Thomas, Vinod and Wang, Yan and Fan, Xibo},
  journal={Journal of Educational Planning and Administration},
  volume={17},
  number={1},
  year={2003}
}

@article{de2007income,
  title={Income inequality measures},
  author={De Maio, Fernando G},
  journal={Journal of Epidemiology \& Community Health},
  volume={61},
  number={10},
  pages={849--852},
  year={2007},
  publisher={BMJ Publishing Group Ltd}
}

@article{halford2024hotspot,
  title={Do hotspot policing interventions against optimal foragers cause crime displacement?},
  author={Halford, Eric and Giannoulis, Mary and Condon, Camie and Keningale, Paige},
  journal={International Journal of Law, Crime and Justice},
  volume={77},
  pages={100654},
  year={2024},
  publisher={Elsevier}
}

@misc{predictive_policing_code,
  author = {Samin Semsar},
  title  = {Predictive Policing Project Code and Data},
  howpublished = {\url{https://github.com/saminsemsar/Data_Analysis_Portfolio/tree/main/PredictivePolicing}},
  year   = {2025}
}

@article{ferguson2012predictive,
  title={Predictive policing and reasonable suspicion},
  author={Ferguson, Andrew Guthrie},
  journal={Emory LJ},
  volume={62},
  pages={259},
  year={2012},
  publisher={HeinOnline}
}

\end{document}